\documentclass{article}
\usepackage{iclr2027_conference,times}

\usepackage{amsmath,amsfonts,bm}

\def\eqref#1{equation~\ref{#1}}

\def\1{\bm{1}}

\DeclareMathAlphabet{\mathsfit}{\encodingdefault}{\sfdefault}{m}{sl}
\SetMathAlphabet{\mathsfit}{bold}{\encodingdefault}{\sfdefault}{bx}{n}

\usepackage{amsmath,amssymb}
\usepackage{booktabs}
\usepackage{graphicx}
\usepackage{pdfpages}
\usepackage{float}
\usepackage{capt-of}
\usepackage{placeins}
\usepackage{needspace}
\usepackage{wrapfig}
\usepackage{xcolor}
\usepackage{colortbl}
\usepackage{tabularx}
\usepackage{etoc}
\usepackage{titletoc}
\usepackage{tikz}
\usetikzlibrary{arrows.meta,positioning,shapes.geometric}
\definecolor{metricgroup}{HTML}{EEE6F4}
\definecolor{metricfocus}{HTML}{E5F1FF}
\definecolor{metricfocuspurple}{HTML}{F3ECFB}
\definecolor{metricprojection}{HTML}{F3F3F3}
\usepackage{hyperref}
\usepackage{url}
\usepackage{marvosym}

\title{{\Large Physical Canvas: A Vision Tactile World Model\\
for Robot Policy Evaluation and Learning}}

\author{%
\begin{minipage}{\dimexpr\textwidth-2\tabcolsep\relax}
\vspace*{0.14in}
\raggedright\normalfont
\textbf{Zebin Xing}$^{1,2,3,*}$ \quad
\textbf{Yuhang Zheng}$^{4,*}$ \quad
\textbf{Yupeng Zheng}$^{1,2,3,*}$ \quad
\textbf{Yichen Zhang}$^{1,3,*}$ \\
\textbf{Zichen Li}$^{3,5}$ \quad
\textbf{Shangqing Zhou}$^{1,2}$ \quad
\textbf{Yujie Zang}$^{4}$ \quad
\textbf{Yuchen Liu}$^{6}$ \\
\textbf{Zhongpu Xia}$^{2}$ \quad
\textbf{Haoran Li}$^{1}$ \quad
\textbf{Qichao Zhang}$^{1}$\textsuperscript{\Letter} \\[0.55em]
\small
$^{1}$Institute of Automation, Chinese Academy of Sciences \quad
$^{2}$Anyverse Robotics \quad
$^{3}$Open-Phoenix Team \\
$^{4}$National University of Singapore \quad
$^{5}$Xi'an Jiaotong University \quad
$^{6}$Beijing Institute of Technology \\[0.35em]
\footnotesize
\textsuperscript{*}\,Equal contribution. \quad
\textsuperscript{\Letter}\,Corresponding author.
\end{minipage}%
}

\newcommand{\method}{\textsc{PhysCanvas}}
\newcounter{physcanvasalgorithm}

\iclrfinalcopy

\begin{document}
\raggedbottom
\maketitle
\lhead{Under review}
\rhead{}
\renewcommand{\headrulewidth}{0.4pt}

\begin{abstract}
Evaluating and post-training robot policies through physical interaction is costly and difficult to scale, motivating action-conditioned world models as interactive simulators. However, visually plausible predictions can misrepresent contact dynamics, providing unreliable feedback for policy evaluation and learning. We introduce \method{}, a state-consistent visuo-tactile world model for contact-rich robot manipulation. \method{} unifies multi-view RGB, tactile images, marker displacement, task state, and actions within a unified vision space. All modalities share a video VAE and pretrained generative backbone, allowing optical touch to reuse visual and temporal priors. Motion images derived through forward kinematics and camera projection provide spatially aligned action conditioning, while an initial RGB anchor and updated reference observations guide sequential generation. Our consistency-aware post-training uses the frozen Scene Reward Model, Modality Reward Model, and State Reward Model to supervise scene preservation, visual--tactile compatibility, and agreement with real task outcomes. The resulting world model supports closed-loop policy rollouts for evaluation and provides imagined interactions and rewards for GRPO policy updates. Across four real-world manipulation tasks, when policies are rolled out in the world model, State Reward Model supervision increases the arithmetic mean of task-wise Pearson correlations between world-model and real-world success rates from $0.602$ to $0.885$. Policy post-training in the reward-guided world model further increases $\pi_{0.5}$'s mean real-world success rate from a $59.1\%$ SFT baseline to $70.8\%$. These results demonstrate the value of unified multimodal modeling and consistency-aware post-training for robot policy evaluation and learning. \textbf{Project page:} \url{https://zebinx.github.io/PhysCanvas_HomePage/}.
\end{abstract}

\section{Introduction}
\label{sec:introduction}

Robot world models predict how environments evolve in response to actions, offering low-cost, reproducible interactions for policy evaluation and learning~\citep{ha2018worldmodels,wu2023daydreamer,wang2026interactive}. Contact-rich manipulation requires both visual observations of scene dynamics and tactile feedback about contact changes~\citep{yuan2017gelsight,lambeta2020digit,lee2020making,guzey2024seetotouch}. A world model that captures these complementary signals and their associated task outcomes could reduce the need for repeated physical trials.

Recent action-conditioned video models combine pretrained generative priors with action representations to support interactive prediction, policy evaluation, and learning in imagination~\citep{wu2026imac,wang2026interactive,yang2026rise}. Existing visuo-tactile models, however, typically require a dedicated tactile encoder and a custom interface for injecting tactile features into a general backbone~\citep{zheng2026omnivta,ma2026feelworld}. Adapting to new tactile data or sensor configurations may therefore require retraining, limiting generality and complicating the pipeline. Meanwhile, token-wise reconstruction supervision emphasizes local fidelity and can miss high-level semantics such as task success. Reliable rollouts thus require visual observations, tactile feedback, and predicted task outcomes to describe the consequences of the same action; inconsistencies can mislead policy optimization~\citep{yang2026mirabench,jiang2026wovr}. For example, after an object slips, predicting stable tactile contact or eventual success is semantically inconsistent despite visually plausible frames. The challenge is to build a reusable multimodal model while preserving scene, cross-modal, and task-outcome consistency for policy evaluation and learning.

\vspace{-4pt}
\noindent \textbf{We investigate two questions: (1) Can a simple, general mechanism encode heterogeneous modalities in a shared space? (2) Can high-level semantics be injected into the world model without additional networks?}

\begin{figure}[t]
  \centering
  \setlength{\abovecaptionskip}{4pt}
  \setlength{\belowcaptionskip}{0pt}
  \includegraphics[width=\linewidth]{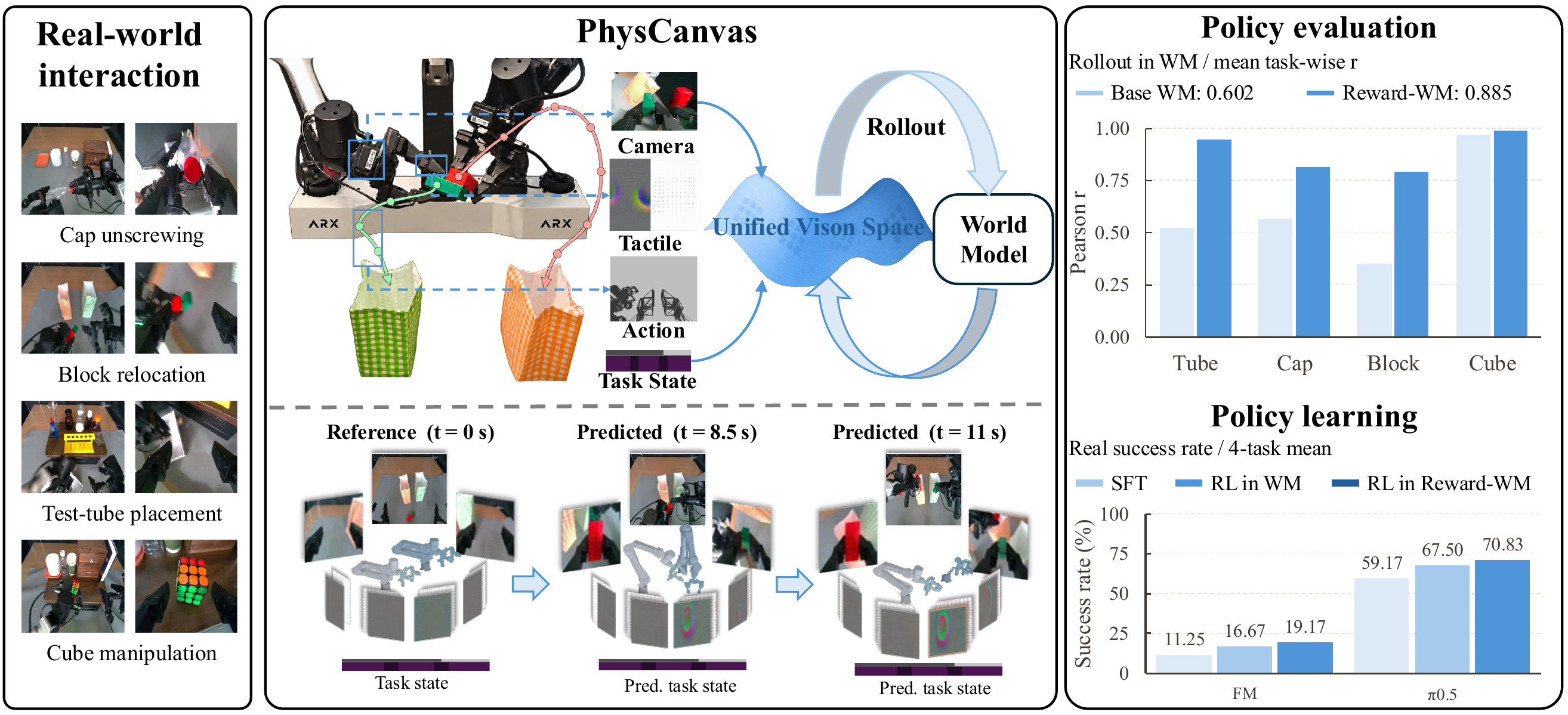}
  \caption{\textbf{PhysCanvas as a multimodal simulator.} Action-conditioned visual--tactile rollouts support policy evaluation and learning.}
  \label{fig:teaser}
  \vspace{-4pt}
\end{figure}

{\pretolerance=10000\hyphenpenalty=10000\exhyphenpenalty=10000\emergencystretch=1em
To address this challenge, we introduce \method{}, a world model with state consistency for visual and tactile prediction, robot policy evaluation, and learning. \method{} places multiple modalities in a shared generative space and uses consistency supervision during post training to preserve the interaction dynamics and task outcomes induced by actions, providing interactive, coherent observations.

Specifically, \method{} places multiview RGB, tactile images, marker displacement, and task state on a unified canvas predicted by a shared video VAE and generative backbone, allowing optical touch to reuse pretrained visual and temporal priors~\citep{wan2025}. Actions are encoded as spatially aligned motion images through forward kinematics and camera projection~\citep{wu2026imac}, while an initial RGB anchor and updated references guide sequential generation. During post training, the frozen Scene Reward Model, Modality Reward Model, and State Reward Model enforce scene preservation, visual and tactile compatibility, and agreement with real outcomes, including both success and failure. The resulting model supports evaluation in a closed loop and supplies imagined interactions and rewards for GRPO policy updates~\citep{shao2024deepseekmath,rlinf}.
}

\begin{figure}[H]
  \centering
  \includegraphics[width=\linewidth]{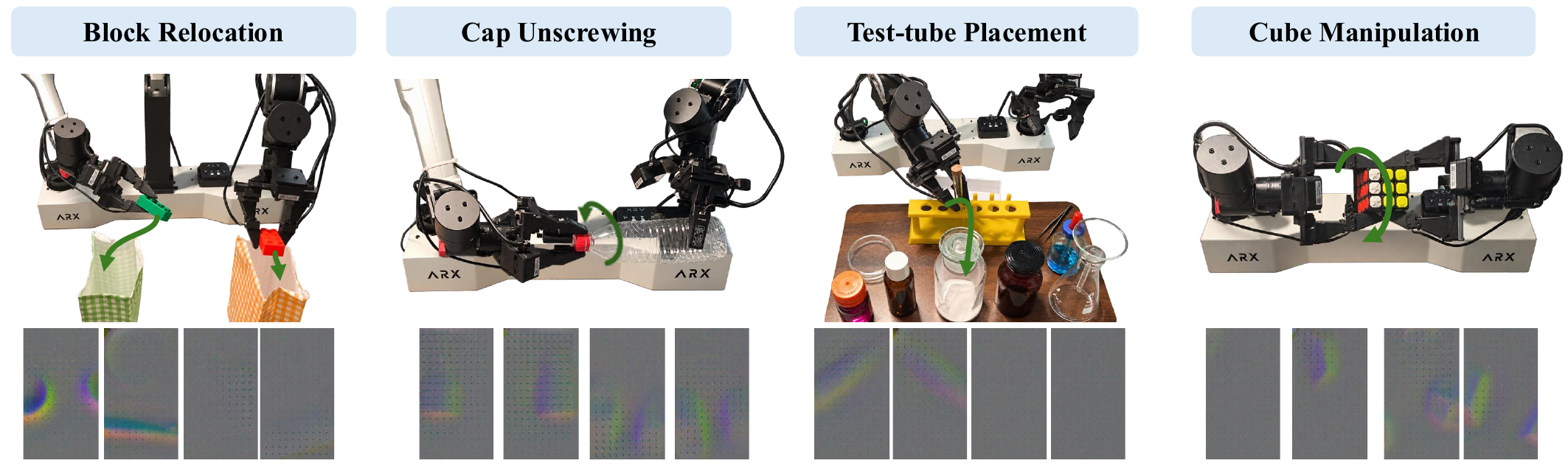}
  \caption{\textbf{Task overview.} Real-world executions and corresponding tactile observations.}
  \label{fig:task_tactile_overview}
\end{figure}

We evaluate \method{} on four real-world manipulation tasks: test-tube placement, bottle-cap unscrewing, block relocation, and Rubik's Cube manipulation (Figure~\ref{fig:task_tactile_overview}). When policies are rolled out in the world model, post-training improves the Pearson correlation between their world-model and real-world success rates on every task, increasing the arithmetic mean of task-wise correlations from $0.602$ to $0.885$. Each task includes nine evaluation batches spanning ACT, FM, and $\pi_{0.5}$~\citep{zhao2023act,lipman2023flowmatching,black2025pi05}; the batches are evaluated at different times, and each task receives equal weight in the summary. Policy optimization through reward-guided imagined interactions further improves the mean real-world success rate of $\pi_{0.5}$ by $11.7$ percentage points over SFT, from $59.1\%$ to $70.8\%$ across the four tasks. These results demonstrate utility for reliable evaluation and policy improvement.

Our contributions are threefold: (1) a unified canvas space that jointly models vision, touch, task state, and actions using a shared video generation framework; (2) consistency-aware post-training that constrains generated interactions through scene, modality, and task-state supervision; and (3) a framework for policy evaluation and learning in imagination, evaluated on real robots.

\section{Related Work}
\label{sec:related_work}

\subsection{Action-Conditioned World Models for Robotics}
Action-conditioned video models have advanced robot planning and interactive prediction~\citep{finn2017visualforesight,yang2024unisim,wu2024ivideogpt,wu2023daydreamer,micheli2023iris,bruce2024genie}. Pretrained video priors~\citep{ho2022videodiffusion,blattmann2023alignlatents,wan2025} and embodied action conditioning~\citep{jiang2025evac,wu2026imac} improve controllability, while unified architectures connect video prediction with policy modeling~\citep{bi2026motus,kim2026cosmospolicy}. Extending these visual simulators to contact-rich manipulation also requires predicting the tactile observations on which policies act.

Imagined rollouts also support policy evaluation and learning~\citep{ha2018worldmodels,hafner2020dreamer,hansen2024tdmpc2,hafner2025dreamerv3,wang2026interactive,jiang2026wovr}, making reliable action consequences and task outcomes essential. Recent work studies long horizon consistency and simulator validity~\citep{gigaworld2026,yang2026mirabench}, or improves generation through reward based optimization~\citep{bardhan2026persistent,wang2025alignavideo}. \method{} brings these requirements to joint visual and tactile prediction through scene and task state consistency supervision, and evaluates generated rollouts against real policy outcomes.

\subsection{Visuo-Tactile World Modeling}
{\pretolerance=10000\hyphenpenalty=10000\exhyphenpenalty=10000\emergencystretch=1em
Touch exposes contact, deformation, and slip that can be ambiguous in RGB, motivating joint visual and tactile prediction~\citep{yuan2017gelsight,lambeta2020digit,lee2020making,guzey2024seetotouch}. OmniVTA~\citep{zheng2026omnivta} combines two stream dynamics with contact aware action generation; FeelWorld~\citep{ma2026feelworld} predicts contact and slip states through contact gated attention. DexTouch-WM~\citep{qin2026dextouch} couples video and tactile experts, whereas DexTacWAM~\citep{yuan2026dextacwam} integrates tactile latents specific to each finger into video diffusion. TACO~\citep{liu2026taco} uses generated visual and tactile corrections for policy post training. Rather than add a separate tactile predictor, \method{} renders tactile appearance and marker displacement as regions of the same visual canvas as RGB and task status. The shared visual VAE and video denoiser then model aligned sensory futures, with explicit supervision of visual and tactile agreement.\par}

\section{Method}
\label{sec:method}

We first define the prediction problem (Section~\ref{sec:problem}) and introduce our unified multimodal representation (Section~\ref{sec:canvas}). We then describe world-model post-training (Section~\ref{sec:posttraining}) and its use for policy evaluation and policy post-training (Section~\ref{sec:simulator}).

\begin{figure}[t]
  \centering
  \includegraphics[width=\linewidth]{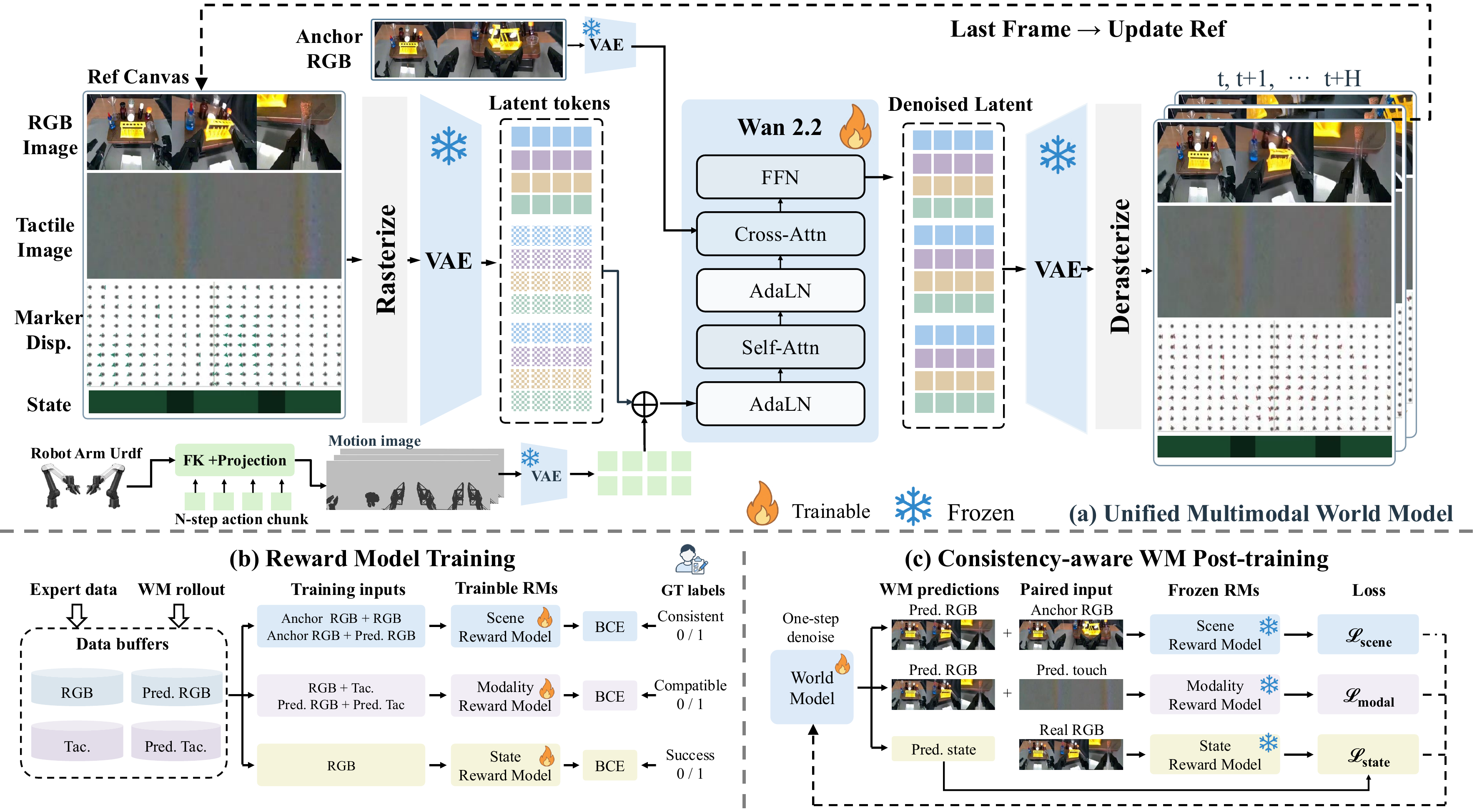}
  \caption{\textbf{Overview of \method.} A unified canvas supports action-conditioned multimodal generation, with scene, state, and modality consistency guiding post-training.}
  \label{fig:method}
\end{figure}

\subsection{Problem Formulation}
\label{sec:problem}

We consider an offline dataset $\mathcal{D}=\{\tau_n\}_{n=1}^{N}$ of robot trajectories. Each trajectory is a time-aligned sequence of observation--action pairs,
\begin{equation}
    \tau_n=\left((\mathbf{o}_0,\mathbf{a}_0),\ldots,
    (\mathbf{o}_{T_n-1},\mathbf{a}_{T_n-1})\right),
    \label{eq:dataset}
\end{equation}
where $\mathbf{o}_t=(\mathbf{o}^{\mathrm{cam}}_t,\mathbf{o}^{\mathrm{tac}}_t,\mathbf{s}_t)$ comprises synchronized multi-view camera observations, tactile sensor measurements, and task-state information $\mathbf{s}_t$. The action condition $\mathbf{a}_t$ is the target dual-arm joint configuration, including the gripper state, associated with time $t$. Thus, $\mathbf{a}_{t:t+H}$ contains the current configuration and $H$ future target configurations, for a total of $H+1$ states, and describes the commanded robot motion over the prediction horizon.

We build our world model from a pretrained image-to-video (I2V) model and extend its visual and temporal priors to action-conditioned multimodal prediction. Given the fixed episode-initial RGB anchor $\mathbf{o}^{\mathrm{cam}}_0$, current multimodal reference $\mathbf{o}_t$, and target joint-configuration sequence $\mathbf{a}_{t:t+H}$, the model predicts the next $H$ frames:
\begin{equation}
    p_{\theta}\!\left(\mathbf{o}_{t+1:t+H}\mid
    \mathbf{o}^{\mathrm{cam}}_0,\mathbf{o}_t,\mathbf{a}_{t:t+H}\right).
    \label{eq:world_model}
\end{equation}

\subsection{Unified Multimodal Canvas}
\label{sec:canvas}

\begin{wrapfigure}[12]{r}{0.46\linewidth}
  \vspace{-9pt}
  \centering
  \includegraphics[width=\linewidth]{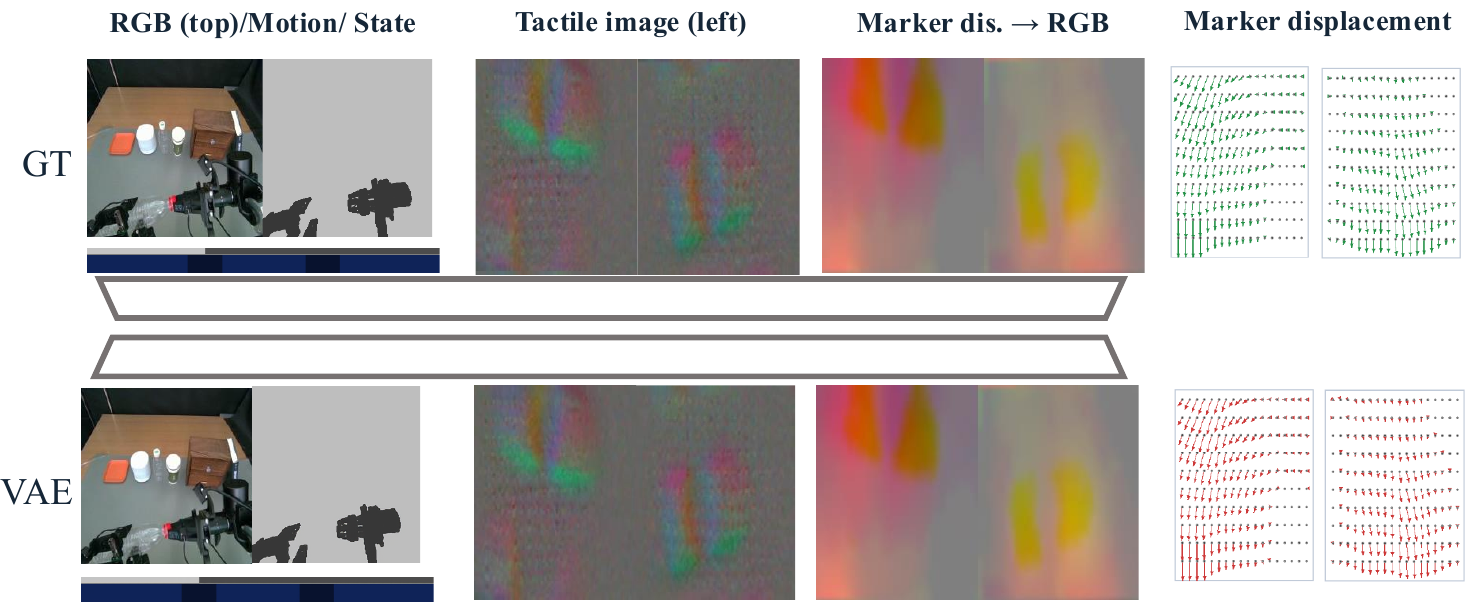}
  \caption{\small\textbf{Multimodal VAE reconstruction.} Inputs (top) and reconstructions (bottom).}
  \label{fig:tactile_vae}
  \vspace{-8pt}
\end{wrapfigure}

\paragraph{Image-native modality rendering.}
\begin{wraptable}{r}{0.47\linewidth}
  \vspace{-10pt}
  \centering
  \caption{\small Tactile image reconstruction on the USB 3D-pose dataset released with UniT~\citep{xu2025unit}. $\dagger$: zero-shot, without tactile fine-tuning.}
  \label{tab:tactile_vae}
  \footnotesize
  \setlength{\tabcolsep}{3pt}
  \begin{tabular}{@{}lrrr@{}}
    \toprule
    \rowcolor{metricgroup}
    Model & PSNR$\uparrow$ & SSIM$\uparrow$ & LPIPS$\downarrow$ \\
    \midrule
    SD ft-MSE$^{\dagger}$ & 30.73 & 0.9020 & 0.0600 \\
    SDXL fp16-fix$^{\dagger}$ & 32.28 & 0.9226 & 0.0399 \\
    UniT & 31.72 & \textbf{0.9613} & 0.0210 \\
    \rowcolor{metricfocus}
    Wan VAE$^{\dagger}$ & \textbf{35.51} & 0.9539 & \textbf{0.0139} \\
    \bottomrule
  \end{tabular}
  \vspace{-8pt}
\end{wraptable}
Existing visuo-tactile models commonly train a dedicated tactile VAE or encoder to obtain tactile representations~\citep{xu2025unit,zheng2026omnivta}. While effective within their training domains, their generality is constrained by the limited scale and diversity of available tactile data. Yet optical tactile images, like natural images, exhibit local spatial correlations and translation equivariance. This motivates a simpler question: can a visual VAE pretrained on large-scale image and video data directly encode tactile observations?
To test this premise, Table~\ref{tab:tactile_vae} compares the tactile-specific UniT VAE~\citep{xu2025unit} with the pretrained SD~\citep{rombach2022ldm}, SDXL~\citep{podell2024sdxl}, and Wan~\citep{wan2025} visual VAEs on the USB 3D-pose tactile dataset released with UniT. The visual VAEs are evaluated without tactile fine-tuning. The Wan VAE achieves higher PSNR (35.51 vs.\ 31.72) and lower LPIPS (0.0139 vs.\ 0.0210) than the tactile-specific UniT VAE, suggesting that its visual representation transfers to optical touch. Figure~\ref{fig:tactile_vae} complements this comparison with VAE reconstructions of RGB, tactile images, RGB-encoded marker displacement, and task state.

Building on this evidence, we adopt the pretrained Wan image-to-video (I2V) model~\citep{wan2025} to exploit its visual and temporal priors for joint multimodal prediction. \method{} organizes camera RGB $\mathbf{I}^{\mathrm{cam}}_t$, tactile images $\mathbf{I}^{\mathrm{tac}}_t$, marker-displacement fields $\mathbf{M}^{\mathrm{tac}}_t$, and task state $\mathbf{s}_t$ in a shared RGB canvas, allowing all modalities to share the visual VAE and video-generation backbone. Actions are likewise rendered as motion images in this visual space and drive its temporal evolution. Tactile images capture contact locations and local surface geometry, while marker displacement records each marker's three-dimensional offset from its reference configuration, providing cues to contact deformation and relative object motion. We map its $x$, $y$, and $z$ components to R, G, and B channels through $g_{\mathrm{disp}}$, using fixed normalization across training and inference. Task state indicates whether the task is complete. Predicting it serves as an auxiliary objective that teaches the world model high-level task semantics; the predicted state is not used by downstream policies. Color denotes the task and texture the outcome; Figure~\ref{fig:task_state_canvas} shows all encodings. RGB and tactile images are inserted directly, whereas marker displacement and state use these image mappings. Camera views are ordered horizontally and modality tiles are stacked vertically:
\begin{equation}
    \mathbf{x}_t = \operatorname{Rasterize}(\mathbf{o}_t) = \operatorname{Concat}_{v}\!\left[\mathbf{I}^{\mathrm{cam}}_t,\mathbf{I}^{\mathrm{tac}}_t,g_{\mathrm{disp}}(\mathbf{M}^{\mathrm{tac}}_t),g_{\mathrm{state}}(\mathbf{s}_t)\right].
    \label{eq:canvas}
\end{equation}
This layout preserves within-modality spatial structure while exposing all modalities to shared spatiotemporal self-attention.

\paragraph{Action-conditioned latent dynamics.}
\label{sec:conditioning}
After arranging the prediction modalities on the canvas, we condition its evolution on robot motion. Following iMaC~\citep{wu2026imac}, we use camera intrinsics and extrinsics together with the robot's URDF to project the arm and gripper configuration into each camera view, producing action images that spatially align robot motion with visual observations.

We perform joint prediction in latent space. The canvas sequence $\mathbf{x}$ and the spatially arranged action-image sequence $\mathbf{m}$ are separately encoded by the video VAE $E$ and tokenized by $P$, giving $\mathbf{z}=P(E(\mathbf{x}))$ and $\mathbf{z}^{a}=P(E(\mathbf{m}))$. The action layout is aligned with the canvas token grid so that conditioning can be added position-wise. At denoising time $\tau$, we fuse action tokens with the noisy canvas tokens as
\begin{equation}
    \widetilde{\mathbf{z}}_{\tau,i}=\mathbf{z}_{\tau,i}+M_i\mathbf{z}^{a}_i,
    \qquad
    M_i=\begin{cases}0,& i\in\mathcal{S}_{\mathrm{state}},\\1,&\text{otherwise},\end{cases}
    \label{eq:action_token_fusion}
\end{equation}
where $i$ indexes aligned spatiotemporal tokens and $\mathcal{S}_{\mathrm{state}}$ contains the task-state token positions. Excluding direct action addition at these positions encourages state prediction to draw on the other modalities through shared self-attention, rather than a dedicated action-to-state input.

The fused tokens are processed by the Wan 2.2 diffusion transformer~\citep{wan2025}. Let $\mathbf z^{\mathrm{anc}}=P(E(\mathbf I^{\mathrm{cam}}_0))$ be the episode-initial RGB anchor, fixed throughout the rollout, and $\mathbf r=P(E(\mathbf x_t))$ the current multimodal reference tokens, updated from the preceding chunk's last frame. The transformer receives $\mathbf r$ as observed reference tokens and $\mathbf z^{\mathrm{anc}}$ through cross-attention. We train the model with flow matching. Using $\tau=0$ for clean data and $\tau=1$ for noise, we interpolate future canvas latents as $\mathbf{z}_\tau=(1-\tau)\mathbf{z}+\tau\epsilon$, with $\epsilon\sim\mathcal{N}(0,I)$ and target velocity $\mathbf{u}_\tau=\epsilon-\mathbf{z}$. The objective is
\begin{equation}
    \mathcal{L}_{\mathrm{FM}}
    = \mathbb{E}_{\mathbf{z},\epsilon,\tau}
    \left[\left\|v_\theta(\widetilde{\mathbf{z}}_\tau,\tau;\mathbf r,\mathbf z^{\mathrm{anc}})-
    \mathbf{u}_\tau\right\|_2^2\right].
    \label{eq:fm}
\end{equation}
The loss supervises future tokens across the generated modalities; observed reference tokens remain conditioning inputs.

\subsection{Consistency-Aware Post-Training}
\label{sec:posttraining}

Closed-loop simulation requires a reliable and consistent world model, beyond pixel-level similarity to observed trajectories. Pixel token wise supervision is insufficient to enforce these requirements; we therefore introduce reward-based post-training to improve scene, state, and modality consistency.

\subsubsection{Reward Models}
\label{sec:reward_models}
We use a separate lightweight reward model for each consistency criterion. Given criterion-specific input $\mathbf{u}_q=h_q(\mathbf{x}_0,\mathbf{x}_{t+h})$, a ResNet-18 encoder~\citep{he2016resnet} $\phi_q$ and decoder $D_q$ produce a logit and score:
\begin{equation}
    R_q(\mathbf{u}_q)=D_q\!\left(\phi_q(\mathbf{u}_q)\right),
    \qquad \rho_q(\mathbf{u}_q)=\sigma\!\left(R_q(\mathbf{u}_q)\right),
    \label{eq:reward_architecture}
\end{equation}
where $D_q$ may be an MLP or Transformer decoder and $\rho_q\in[0,1]$. Each model is trained beforehand and remains frozen while supervising the generator.

\paragraph{Scene consistency.}
The Scene Reward Model takes an episode-initial RGB reference and a future RGB frame, assigning $(1)$ to preserved scene identity and objects under valid motion and $(0)$ to unintended changes. For training, we construct a class-balanced dataset using ground-truth observations and scene-consistent world-model generations as positive examples, and scene-inconsistent generations as negative examples. Each example is paired with its episode-initial RGB reference.

\paragraph{State consistency.}
\label{sec:state}
The State Reward Model takes a future RGB observation and predicts whether the task's success condition holds ($1$) or remains unsatisfied ($0$). Its judgments provide labels for the canvas state region, whose color identifies the task and texture encodes completion. We train it exclusively on real trajectories, which naturally contain abundant negative examples: frames before task completion are labeled $(0)$, while frames from completion onward are labeled $(1)$.

\paragraph{Modality consistency.}
The Modality Reward Model takes paired RGB, tactile-image, and marker-displacement regions, assigning $(1)$ when their contact evidence is compatible and $(0)$ when it is mismatched. Negative training pairs are constructed by keeping the RGB observation fixed and replacing the paired tactile observations with a different tactile state, such as one with a different contact texture. This supervises cross-modal agreement rather than the presence of contact.

\subsubsection{Reward-Guided World-Model Post-Training}
\label{sec:reward_posttraining}
Reward supervision operates in image space. At a sampled diffusion time $\tau$, we recover a clean canvas estimate from the predicted velocity and decode it with the video VAE:
\begin{equation}
    \widehat{\mathbf{z}}=\mathbf{z}_{\tau}-\tau v_{\theta}(\widetilde{\mathbf{z}}_{\tau},\tau;\mathbf r,\mathbf z^{\mathrm{anc}}),
    \qquad
    \widehat{\mathbf{x}}=D\!\left(P^{-1}(\widehat{\mathbf{z}})\right),
    \label{eq:reward_decode}
\end{equation}
where $P^{-1}$ restores the token sequence to the VAE latent grid and $D$ is the decoder.

\paragraph{State supervision.}
The world model should reproduce the observed task outcome, whether successful or unsuccessful. The frozen State Reward Model evaluates the generated RGB observation and produces a success logit $\widehat{\ell}_{\mathrm{state}}$. We obtain the binary target $y_{\mathrm{state}}$ by thresholding the same model's success score on the corresponding ground-truth RGB observation. We then minimize binary cross-entropy with logit temperature $T_{\mathrm{state}}$:
\begin{equation}
    \mathcal{L}_{\mathrm{state}}
    =\mathbb{E}\!\left[\operatorname{BCEWithLogits}
    \!\left(\widehat{\ell}_{\mathrm{state}}/T_{\mathrm{state}},\ y_{\mathrm{state}}\right)\right].
    \label{eq:state_reward_loss}
\end{equation}
This preserves both successful and unsuccessful outcomes rather than forcing every prediction toward success.

\paragraph{Scene and modality supervision.}
Unlike task success, scene and cross-modal consistency are desirable for every prediction. Following the direct reward-maximization objective of Align-A-Video~\citep{wang2025alignavideo}, we optimize the decoded canvas to increase the scores assigned by the frozen reward models:
\begin{equation}
    \mathcal{L}_q=-\mathbb{E}\!\left[\rho_q\!\left(h_q(\mathbf{x}_0,\widehat{\mathbf{x}})\right)\right],
    \qquad
    q\in\{\mathrm{scene},\mathrm{modality}\}.
    \label{eq:consistency_reward_loss}
\end{equation}
Here $h_q$ selects the inputs required by each reward model, and $\rho_q$ is its consistency score in $[0,1]$. Gradients propagate through the frozen reward models and VAE decoder to update the world model. During reward-guided post-training, we retain the first-stage flow-matching MSE loss to preserve multimodal generation fidelity.

\subsection{Policy Evaluation and Policy Post-Training}
\label{sec:simulator}

\paragraph{Policy evaluation.}
\method{} provides a multimodal environment for evaluating robot policies without executing every trial on hardware. Starting from a real initial observation, each policy is rolled out in the world model in a closed loop: the policy consumes the generated observation and issues the next action, and the world model returns its predicted visual, tactile, and task-state consequences. Repeating this process produces imagined policy trajectories under controlled initial conditions. We assess simulator reliability by comparing the success rates of these world-model rollouts with those of the corresponding real-world policy executions under the same success criteria.

\paragraph{Policy post-training.}
Beyond evaluation, \method{} supports policy post-training without collecting additional robot trajectories. From real initial observations, a policy explores alternative behaviors through imagined closed-loop interactions, while the frozen State Reward Model scores their task outcomes. Group-relative policy optimization~\citep{shao2024deepseekmath} converts differences among these outcomes into learning signals for the policy. Throughout this process, the world model and reward model remain fixed, so only the policy is updated from generated experience.

\setlength{\textfloatsep}{10pt plus 2pt minus 2pt}
\setlength{\intextsep}{8pt plus 2pt minus 2pt}
\setlength{\floatsep}{8pt plus 2pt minus 2pt}
\section{Experiments}
\label{sec:experiments}

We evaluate action-conditioned multimodal prediction, the effect of consistency-aware post-training, and the use of imagined rollouts for policy evaluation and learning.

\begin{table}[!t]
\centering\fontsize{7}{8.4}\selectfont
\setlength{\tabcolsep}{2pt}\renewcommand{\arraystretch}{1.15}
\caption{\textbf{RGB comparison.} Visual metrics (left), consistency scores and tracking errors (right).}
\label{tab:baselines}
\begin{tabularx}{\linewidth}{@{}l>{\raggedleft\arraybackslash}X>{\raggedleft\arraybackslash}X>{\raggedleft\arraybackslash}X>{\raggedleft\arraybackslash}X>{\raggedleft\arraybackslash}X!{\vrule width 0.6pt}>{\raggedleft\arraybackslash}X>{\raggedleft\arraybackslash}X>{\raggedleft\arraybackslash}X>{\raggedleft\arraybackslash}X>{\raggedleft\arraybackslash}X@{}}
\toprule
\rowcolor{metricgroup}
Model & \multicolumn{5}{c}{Conventional RGB metrics} & \multicolumn{5}{c}{VBench-derived / control diagnostics}\\
\cmidrule(lr){2-6}\cmidrule(lr){7-11}
 & PSNR$\uparrow$ & SSIM$\uparrow$ & LPIPS$\downarrow$ & FID$\downarrow$ & FVD$\downarrow$ & Subject$\uparrow$ & Bg.$\uparrow$ & Smooth$\uparrow$ & EEF (px)$\downarrow$ & Object (px)$\downarrow$\\
\midrule
EVAC & 16.439 & 0.528 & 0.258 & 26.512 & 165.705 & 79.223 & 90.456 & 97.731 & 1.964 & 2.150\\
Open-Sora & 17.916 & 0.599 & 0.228 & 34.762 & 183.860 & 78.519 & 90.234 & 98.157 & 1.425 & 1.426\\
WoVR & 17.278 & 0.5972 & 0.203 & 26.086 & 138.243 & 80.247 & \textbf{91.082} & 98.059 & 2.792 & 1.164\\
\rowcolor{metricfocus}
\method{} & \textbf{21.172} & \textbf{0.735} & \textbf{0.121} & \textbf{21.459} & \textbf{116.274} & \textbf{80.349} & 90.565 & \textbf{98.301} & \textbf{0.685} & \textbf{1.150}\\
\bottomrule
\end{tabularx}
\end{table}

\subsection{Experimental Setup}
\label{sec:experimental_setup}

\paragraph{Platform and data.}
We use an ARX AC One dual-arm robot with three cameras and four fingertip Xense tactile sensors (Appendix~\ref{app:hardware}). Our study covers four tasks---test-tube placement, bottle-cap unscrewing, block relocation, and Rubik's Cube manipulation---spanning handover, constrained contact, and object rotation.

\paragraph{Models and training.}
We initialize \method{} from Wan~2.2 and train the unified canvas on mixed-task clips. Training uses four recursive chunks of eight future frames each. Unless otherwise specified, models are trained for 300 epochs. Reward models use ResNet-18 encoders with criterion-specific decoders. We evaluate ACT, FM, and $\pi_{0.5}$ with visual and tactile inputs, and post-train FM and $\pi_{0.5}$.

\paragraph{Baselines and evaluation.}
We compare with EVAC~\citep{jiang2025evac}, Open-Sora~\citep{zheng2024opensora}, and WoVR~\citep{jiang2026wovr}. RGB prediction is evaluated with standard reconstruction and distributional metrics (PSNR, SSIM, LPIPS, FID, and FVD), VBench consistency scores (Subject, Bg., and Smooth)~\citep{huang2024vbench}, and EEF/object tracking errors. For touch, we report tactile-image PSNR, SSIM, LPIPS, Contact MAE, Onset/Offset, and Event F1, together with tactile-flow Active MAE, Idle FP-MAE, and Amp. MAE. Appendix~\ref{app:metric_definitions} gives the definitions and aggregation rules.

\subsection{Comparison with Baselines}
\label{sec:baseline_comparison}

\begin{figure}[!t]
  \centering
  \includegraphics[width=\linewidth]{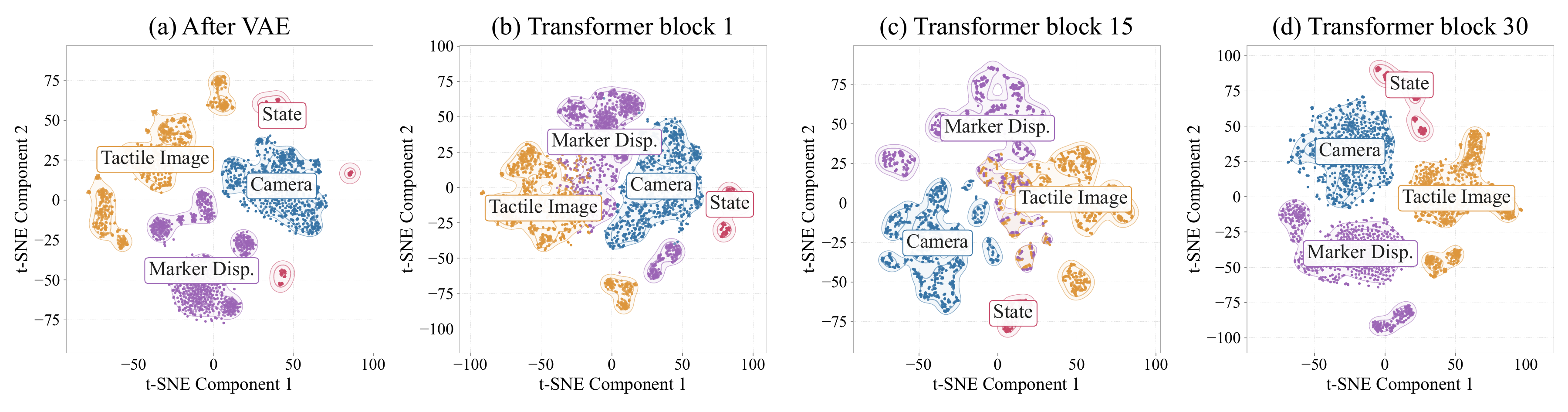}
  \caption{\textbf{Evolution of multimodal token distributions.} t-SNE embeddings of multimodal tokens at (a) the VAE output and after Transformer blocks (b) 1, (c) 15, and (d) 30. Colors denote camera RGB, tactile image, marker displacement, and task state.}
  \label{fig:modality_tsne}
\end{figure}

\begin{wrapfigure}{r}{0.46\linewidth}
  \vspace{-10pt}
  \centering
  \includegraphics[width=\linewidth]{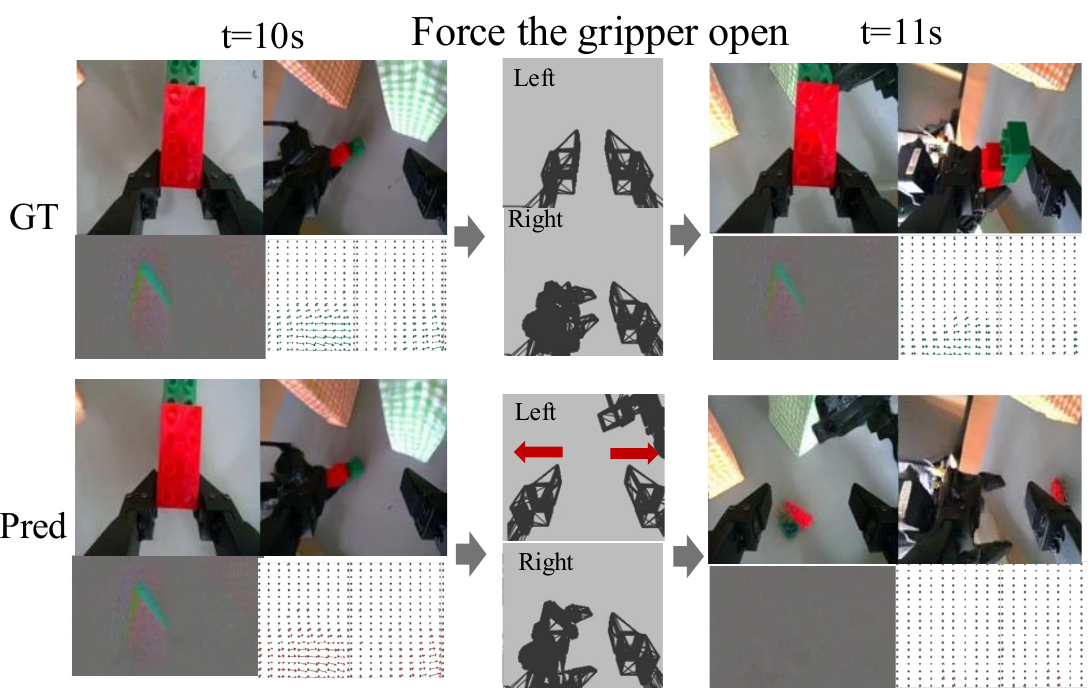}
  \caption{\small\textbf{Action intervention.} Forced-open gripper predicts a falling block.}
  \label{fig:action_ablation}
  \vspace{-8pt}
\end{wrapfigure}

\paragraph{Quantitative comparison.}
Table~\ref{tab:baselines} summarizes RGB prediction results for \method{}, EVAC, Open-Sora, and WoVR. Reconstruction metrics are complemented by temporal-consistency scores and tracking errors to assess visual fidelity and interaction geometry. Overall, \method{} performs strongly across the conventional RGB metrics in Table~\ref{tab:baselines}, achieving the best PSNR, SSIM, LPIPS, FID, and FVD. It has modest advantages on the VBench Subject and Smooth scores, while the larger gain appears in action-sensitive tracking: its EEF error is $0.685$ pixels, $51.9\%$ lower than the next-best baseline, indicating more accurate prediction of gripper position. To examine how \mbox{different} canvas features evolve through the network, Figure~\ref{fig:modality_tsne} uses tSNE to visualize VAE \mbox{outputs} and \mbox{Transformer} value tokens. The visualization includes camera RGB, tactile image, marker \mbox{displacement}, and task state. Representations produced by the VAE form relatively clear \mbox{modality} groups. In the \mbox{Transformer}, value tokens in shallow layers exhibit more local overlap across \mbox{modalities}, whereas deeper value representations become more separable by modality. This \mbox{pattern} \mbox{suggests} that \mbox{interactions} across \mbox{modalities} occur primarily in the shallow blocks.

\FloatBarrier

\subsection{Ablation Studies}
\label{sec:ablation_studies}
\begin{figure}[H]
  \centering
  \setlength{\abovecaptionskip}{4pt}
  \setlength{\belowcaptionskip}{0pt}
  \includegraphics[width=\linewidth]{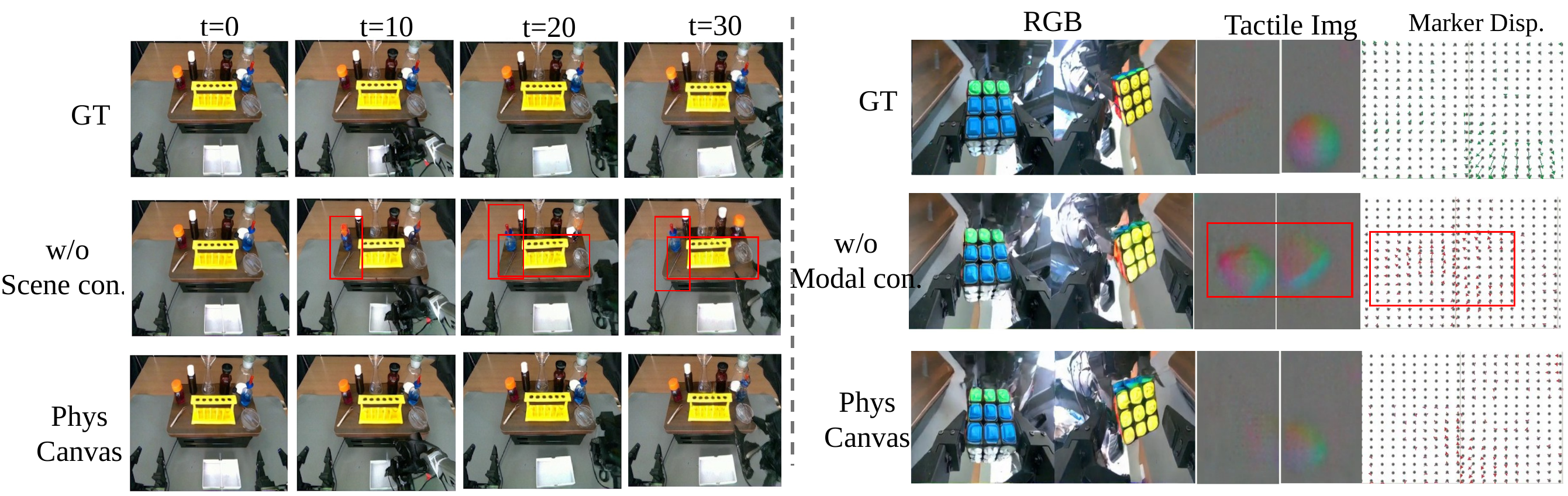}
  \caption{\textbf{Scene and modality consistency.} Left: temporal drift without Scene Reward Model supervision. Right: Rubik's Cube RGB, tactile image, and marker displacement before and after Modality Reward Model supervision; red boxes mark discrepancies.}
  \label{fig:consistency_ablation}
\end{figure}

Table~\ref{tab:ablations} compares two action-injection schemes and three consistency objectives. $\mathcal{M}_0$ injects a global motion embedding with the diffusion timestep through AdaLN~\citep{peebles2023dit}, whereas $\mathcal{M}_1$ uses view-aligned motion images fused position-wise with canvas tokens. Subsequent variants add scene, state, and multimodal consistency supervision.
Compared with $\mathcal{M}_0$, spatially aligned $\mathcal{M}_1$ improves RGB PSNR/SSIM by 23.6\%/18.3\% and lowers LPIPS, FVD, and EEF error by 58.3\%, 21.7\%, and 32.6\%. Tactile PSNR, LPIPS, Contact MAE, and marker-displacement MAEs improve, while a few other metrics worsen.
Scene consistency ($\mathcal{M}_1\!\rightarrow\!\mathcal{M}_2$) accompanies a 14.2\% reduction in RGB LPIPS and a 10.4\% reduction in FVD, with higher subject and background consistency, matching the goal of preserving scene appearance. With state supervision, $\mathcal{M}_3$ further reduces FID by 26.4\% and FVD by 12.5\%, while SSIM increases from 0.693 to 0.720. Multimodal consistency ($\mathcal{M}_3\!\rightarrow\!\mathcal{M}_4$) lowers Contact MAE (0.02491 to 0.02278) and all three marker-displacement MAEs (Active: 4.666 to 4.083; Idle FP: 0.646 to 0.577; amplitude: 7.854 to 6.654), while Event F1 dips from 0.4550 to 0.4455. These metrics assess tactile fidelity; Figure~\ref{fig:consistency_ablation} (right) shows paired visual--tactile results.
\Needspace{30\baselineskip}
\begin{table}[H]
\centering\fontsize{7.2}{8.6}\selectfont
\setlength{\tabcolsep}{2pt}\renewcommand{\arraystretch}{1.12}
\caption{\textbf{Multimodal ablations.} $\mathcal{M}_0$: AdaLN action injection; $\mathcal{M}_1$: spatial motion-image injection; $\mathcal{M}_2$: $\mathcal{M}_1$ + scene consistency; $\mathcal{M}_3$: $\mathcal{M}_2$ + state supervision; $\mathcal{M}_4$: $\mathcal{M}_3$ + multimodal consistency.}
\label{tab:ablations}
\textbf{RGB}\par\vspace{2pt}
\begin{tabularx}{\linewidth}{@{}l*{10}{>{\raggedleft\arraybackslash}X}@{}}
\toprule
\rowcolor{metricgroup}
Model & \multicolumn{5}{c}{Reconstruction} & \multicolumn{3}{c}{VBench scores} & \multicolumn{2}{c}{Tracking error}\\
\cmidrule(lr){2-6}\cmidrule(lr){7-9}\cmidrule(lr){10-11}
 & PSNR$\uparrow$ & SSIM$\uparrow$ & LPIPS$\downarrow$ & FID$\downarrow$ & FVD$\downarrow$ & Subject$\uparrow$ & Bg.$\uparrow$ & Smooth$\uparrow$ & EEF (px)$\downarrow$ & Object (px)$\downarrow$\\
\midrule
$\mathcal{M}_0$ & 16.070 & 0.558 & 0.271 & 22.331 & 197.779 & 79.513 & 89.735 & \textbf{98.588} & 1.201 & \textbf{1.017}\\
$\mathcal{M}_1$ & 19.869 & 0.660 & 0.113 & 33.145 & 154.901 & 80.272 & 90.347 & 98.333 & 0.810 & 1.651\\
$\mathcal{M}_2$ & 20.975 & 0.693 & \textbf{0.097} & 30.719 & 138.761 & \textbf{80.544} & 90.526 & 98.312 & 0.719 & 1.527\\
$\mathcal{M}_3$ & 20.699 & 0.720 & 0.132 & 22.619 & 121.484 & 80.220 & 90.549 & 98.334 & 0.783 & 1.323\\
\rowcolor{metricfocus}
$\mathcal{M}_4$ & \textbf{21.172} & \textbf{0.735} & 0.121 & \textbf{21.459} & \textbf{116.274} & 80.349 & \textbf{90.565} & 98.301 & \textbf{0.685} & 1.150\\
\bottomrule
\end{tabularx}

\vspace{5pt}\textbf{Tactile}\par\vspace{2pt}
\begin{tabularx}{\linewidth}{@{}l*{10}{>{\raggedleft\arraybackslash}X}@{}}
\toprule
\rowcolor{metricgroup}
Model & \multicolumn{7}{c}{Tactile Image} & \multicolumn{3}{c}{Marker Disp. (MAEs $\times10^3$)}\\
\cmidrule(lr){2-8}\cmidrule(lr){9-11}
 & PSNR$\uparrow$ & SSIM$\uparrow$ & LPIPS$\downarrow$ & Contact MAE$\downarrow$ & Onset (s)$\downarrow$ & Offset (s)$\downarrow$ & F1$\uparrow$ & Active MAE$\downarrow$ & Idle FP-MAE$\downarrow$ & Amp. MAE$\downarrow$\\
\midrule
$\mathcal{M}_0$ & 36.059 & 0.9330 & 0.1550 & 0.03012 & 1.638 & 1.241 & 0.4970 & 5.313 & 1.135 & 8.433\\
$\mathcal{M}_1$ & 36.292 & 0.9312 & 0.1265 & 0.02609 & 1.182 & 0.987 & 0.3498 & 4.834 & 0.900 & 7.634\\
$\mathcal{M}_2$ & 37.153 & 0.9349 & 0.1139 & 0.02458 & \textbf{0.663} & 0.900 & \textbf{0.5335} & 4.452 & 0.684 & 7.377\\
$\mathcal{M}_3$ & 36.921 & 0.9326 & 0.1221 & 0.02491 & 0.718 & 0.881 & 0.4550 & 4.666 & 0.646 & 7.854\\
\rowcolor{metricfocus}
$\mathcal{M}_4$ & \textbf{37.598} & \textbf{0.9390} & \textbf{0.1032} & \textbf{0.02278} & 0.664 & \textbf{0.604} & 0.4455 & \textbf{4.083} & \textbf{0.577} & \textbf{6.654}\\
\bottomrule
\end{tabularx}
\end{table}

\vspace{-6pt}
\vspace{-4pt}

\Needspace{7\baselineskip}
\paragraph{Visual diagnostics.}
Figures~\ref{fig:action_ablation} and~\ref{fig:consistency_ablation} examine action response and consistency. In Figure~\ref{fig:action_ablation}, forced gripper opening makes the block fall and the tactile response disappear. Figure~\ref{fig:consistency_ablation} (left) shows progressive scene drift without Scene Reward Model supervision, while the full model preserves the layout. On Rubik's Cube rotation (right), Modality Reward Model supervision suppresses diffuse tactile activity and retains the localized response. These examples are qualitative rather than contact-accuracy measurements.

\subsection{PhysCanvas as a Policy Evaluator}
\label{sec:experiment_policy_evaluation}
\begin{figure}[!t]
  \centering
  \includegraphics[width=\linewidth]{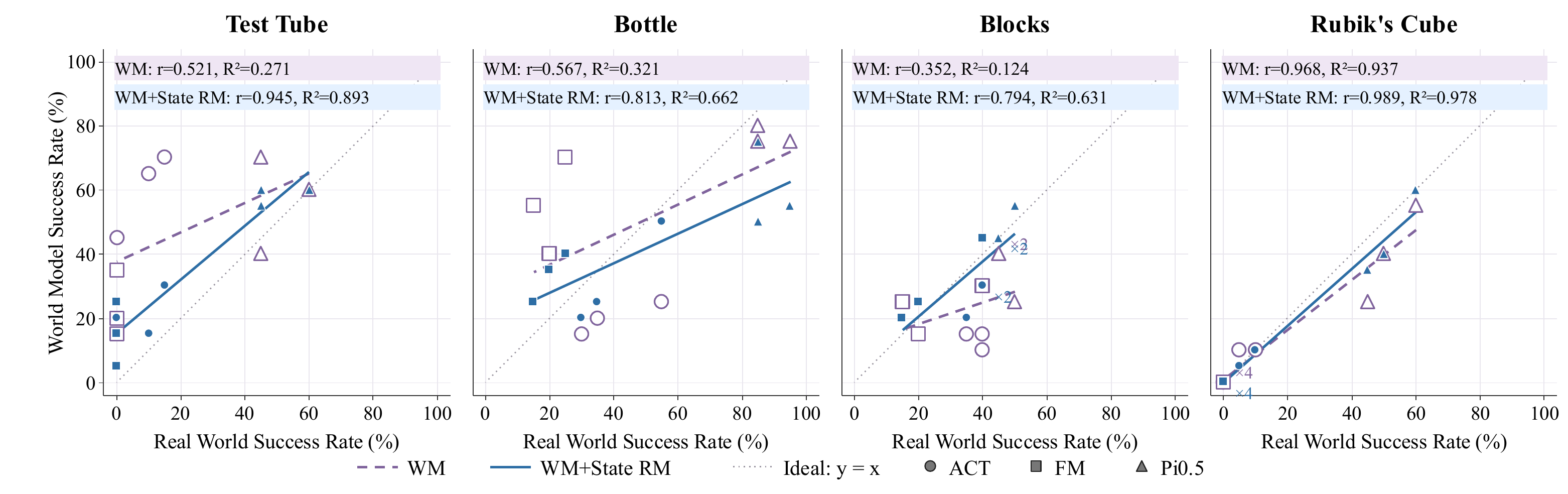}
  \caption{\textbf{Effect of state supervision on policy evaluation.} Each point compares world-model rollout and real-world success rates for one evaluation batch. Dashed blue and solid orange lines are task-wise fits for $\mathcal{M}_2$ and $\mathcal{M}_3$, respectively; the dotted diagonal denotes ideal agreement.}
  \label{fig:policy_evaluation_correlation}
\end{figure}

To evaluate the effect of state supervision on policy evaluation, we roll out ACT, flow matching (FM), and $\pi_{0.5}$ in both $\mathcal{M}_2$ and $\mathcal{M}_3$ across four tasks. For each task--policy pair, evaluation uses three batches of 20 scenes collected at different times. Multiple human experts jointly judge whether each real-world scene satisfies the task-specific success criterion. Figure~\ref{fig:policy_evaluation_correlation} compares the success rates of these world-model rollouts with the corresponding real-world success rates; stronger correlation indicates more reliable estimates across evaluation batches.

\begin{table}[!t]
\centering\fontsize{9}{10.8}
\selectfont
\setlength{\tabcolsep}{1.5pt}\renewcommand{\arraystretch}{1.2}
\caption{\textbf{Real-world policy success rates (\%).} Values are mean $\pm$ population standard deviation over three runs; bold marks the highest nonzero mean per column (including ties).}
\label{tab:policy_learning}
\begin{tabularx}{\linewidth}{@{}l*{8}{>{\centering\arraybackslash}X}@{}}
\toprule
\rowcolor{metricgroup}
Method & \multicolumn{2}{c}{Test Tube} & \multicolumn{2}{c}{Bottle} & \multicolumn{2}{c}{Blocks} & \multicolumn{2}{c}{Rubik's Cube} \\
\cmidrule(lr){2-3}\cmidrule(lr){4-5}\cmidrule(lr){6-7}\cmidrule(lr){8-9}
 & FM & $\pi_{0.5}$ & FM & $\pi_{0.5}$ & FM & $\pi_{0.5}$ & FM & $\pi_{0.5}$ \\
\midrule
SFT & 0.0 $\pm$ 0.0 & 50.0 $\pm$ 7.0 & 20.0 $\pm$ 4.0 & 88.3 $\pm$ 4.7 & {\fontsize{8}{9.6}\selectfont\mbox{25.0 $\pm$ 10.8}} & 46.6 $\pm$ 2.3 & 0.0 $\pm$ 0.0 & 51.6 $\pm$ 6.2 \\
RL in WM & 0.0 $\pm$ 0.0 & \textbf{60.0 $\pm$ 4.0} & 33.3 $\pm$ 2.3 & \textbf{90.0 $\pm$ 0.0} & 33.3 $\pm$ 8.5 & 56.6 $\pm$ 4.7 & 0.0 $\pm$ 0.0 & 63.3 $\pm$ 4.7 \\
\rowcolor{metricfocus}
RL in WM+RM & 0.0 $\pm$ 0.0 & \textbf{60.0 $\pm$ 8.1} & \textbf{35.0 $\pm$ 4.0} & 86.6 $\pm$ 2.3 & \textbf{41.6 $\pm$ 6.2} & \textbf{63.3 $\pm$ 4.7} & 0.0 $\pm$ 0.0 & \textbf{73.3 $\pm$ 4.7} \\
\bottomrule
\end{tabularx}
\vspace{-4pt}
\end{table}

As shown in Figure~\ref{fig:policy_evaluation_correlation}, adding state supervision from $\mathcal{M}_2$ to $\mathcal{M}_3$ improves agreement between world-model rollout and real-world policy outcomes, raising the mean task-wise Pearson correlation from $0.602$ to $0.885$ and the mean $R^2$ from $0.413$ to $0.791$. Gains are largest on Test Tube and Blocks, moderate on Bottle, and minimal on Rubik's Cube, where $\mathcal{M}_2$ is already near ceiling. Since Table~\ref{tab:ablations} shows mixed changes in low-level reconstruction metrics, the outcome gains suggest better task-semantic modeling rather than improved pixel-level reconstruction.

\subsection{PhysCanvas as a Learning Simulator for Policy Improvement}
\label{sec:experiment_policy_learning}
We next ask whether interaction with a visuo-tactile world model can improve policies on the real robot. Starting from supervised fine-tuning (SFT), we optimize FM and $\pi_{0.5}$ with GRPO in \method{} and evaluate the resulting policies on real tasks. In Table~\ref{tab:policy_learning}, RL in WM uses $\mathcal{M}_1$ as the simulator, whereas RL in WM+RM uses the consistency-post-trained $\mathcal{M}_4$; both use the frozen State Reward Model to score task outcomes. We omit ACT because it does not use the diffusion or flow-matching formulation assumed by our GRPO implementation.

RL in $\mathcal{M}_1$ raises mean real-world success from $11.2\%$ to $16.6\%$ for FM and from $59.1\%$ to $67.5\%$ for $\pi_{0.5}$, averaging equally across the four tasks. Using $\mathcal{M}_4$ as the simulator further raises these means to $19.1\%$ and $70.8\%$. For FM, gains occur on Bottle and Blocks, while Test Tube and Rubik's Cube remain at zero. For $\pi_{0.5}$, $\mathcal{M}_4$ yields its largest additional gains over $\mathcal{M}_1$ on Blocks and Rubik's Cube ($6.6$ and $10.0$ percentage points); Bottle decreases from $90.0\%$ to $86.6\%$. Thus, imagined interactions in the consistency-post-trained world model improve average real-world success, although the benefit varies by task.

\section{Conclusion}
We introduced \method{} to study whether heterogeneous modalities can share a simple representation and whether high level task semantics can be injected without an additional prediction network. It places multiview RGB, tactile images, marker displacement, task state, and actions in a unified vision space, while consistency supervision aligns scene, modality, and task outcomes. Experiments across four real world tasks show that task state supervision improves the agreement between world model rollouts and real policy outcomes, while reward guided imagined interactions support policy improvement on real robots. These results demonstrate the value of unified multimodal representations for robot policy evaluation and learning.

\subsection*{AI use statement}
Generative AI tools assisted literature discovery, language editing, and early-draft organization. The authors reviewed and verified all technical content, citations, claims, code, and generated text, and take responsibility for the final manuscript.

\begingroup
\small
\setlength{\bibsep}{0pt plus 0.3ex}
\bibliography{iclr2027_conference}

@inproceedings{bi2026motus,
  title={{Motus}: A Unified Latent Action World Model},
  author={Bi, Hongzhe and Tan, Hengkai and Xie, Shenghao and Wang, Zeyuan and Huang, Shuhe and Liu, Haitian and Zhao, Ruowen and Feng, Yao and Xiang, Chendong and Rong, Yinze and Zhao, Hongyan and Liu, Hanyu and Su, Zhizhong and Ma, Lei and Su, Hang and Zhu, Jun},
  booktitle={Proceedings of the IEEE/CVF Conference on Computer Vision and Pattern Recognition},
  pages={35101--35113},
  year={2026},
  url={https://openaccess.thecvf.com/content/CVPR2026/html/Bi_Motus_A_Unified_Latent_Action_World_Model_CVPR_2026_paper.html}
}

@article{qin2026dextouch,
  title={{DexTouch-WM}: Learning Action-Conditioned Tactile World Models from Human Touch for Dexterous Robot Manipulation},
  author={Qin, Yan and Chen, Yue and Lin, Wenwei and Liu, Shujia and Lyu, Chuqiao and Su, Kailun and Jin, Weiyang and Yu, Chenze and Luo, Ping and Ding, Wenbo and Chen, Tianxing and Xu, Renjing},
  journal={arXiv preprint arXiv:2609.20649},
  year={2026}
}

@inproceedings{kim2026cosmospolicy,
  title={{Cosmos Policy}: Fine-Tuning Video Models for Visuomotor Control and Planning},
  author={Kim, Moo Jin and Gao, Yihuai and Lin, Tsung-Yi and Lin, Yen-Chen and Ge, Yunhao and Lam, Grace and Liang, Percy and Song, Shuran and Liu, Ming-Yu and Finn, Chelsea and Gu, Jinwei},
  booktitle={International Conference on Learning Representations},
  year={2026},
  url={https://proceedings.iclr.cc/paper_files/paper/2026/hash/748becc400a57c0e31cfe6a2e7951467-Abstract-Conference.html}
}

@article{gigaworld2026,
  title={{GigaWorld-1}: A Roadmap to Build World Models for Robot Policy Evaluation},
  author={{GigaWorld Team} and Ma, Angyuan and Wang, Boyuan and Li, Bohan and Ni, Chaojun and others},
  journal={arXiv preprint arXiv:2607.02642},
  year={2026}
}

@inproceedings{finn2017visualforesight,
  title={Deep Visual Foresight for Planning Robot Motion},
  author={Finn, Chelsea and Levine, Sergey},
  booktitle={IEEE International Conference on Robotics and Automation},
  year={2017}
}

@inproceedings{yang2024unisim,
  title={Learning Interactive Real-World Simulators},
  author={Yang, Sherry and Du, Yilun and Ghasemipour, Seyed and Tompson, Jonathan and Kaelbling, Leslie and Schuurmans, Dale and Abbeel, Pieter},
  booktitle={International Conference on Learning Representations},
  year={2024},
  url={https://proceedings.iclr.cc/paper_files/paper/2024/hash/c4d66eae503694424123b93ac0fbaf17-Abstract-Conference.html}
}

@inproceedings{wu2024ivideogpt,
  title={{iVideoGPT}: Interactive {VideoGPTs} are Scalable World Models},
  author={Wu, Jialong and Yin, Shaofeng and Feng, Ningya and He, Xu and Li, Dong and Hao, Jianye and Long, Mingsheng},
  booktitle={Advances in Neural Information Processing Systems},
  year={2024}
}

@article{jiang2025evac,
  title={{EnerVerse-AC}: Envisioning Embodied Environments with Action Condition},
  author={Jiang, Yuxin and Chen, Shengcong and Huang, Siyuan and Chen, Liliang and Zhou, Pengfei and Liao, Yue and He, Xindong and Liu, Chiming and Li, Hongsheng and Yao, Maoqing and Ren, Guanghui},
  journal={arXiv preprint arXiv:2505.09723},
  year={2025}
}

@article{jiang2026wovr,
  title={{WoVR}: World Models as Reliable Simulators for Post-Training {VLA} Policies with {RL}},
  author={Jiang, Zhennan and Zhou, Shangqing and Jiang, Yutong and Huang, Zefang and Wei, Mingjie and Chen, Yuhui and Zhou, Tianxing and Guo, Zhen and Lin, Hao and Zhang, Quanlu and Wang, Yu and Li, Haoran and Yu, Chao and Zhao, Dongbin},
  journal={arXiv preprint arXiv:2602.13977},
  year={2026}
}

@article{wu2026imac,
  title={{iMaC}: Translating Actions into Motion and Contact Images for Embodied World Models},
  author={Wu, Zhenyu and Xu, Xiuwei and Zhou, Yukun and Li, Yifan and Deng, Qiuping and Wang, Xiaofeng and Zhu, Zheng and Yu, Bingyao and Wang, Ziwei and Lu, Jiwen and Yan, Haibin},
  journal={arXiv preprint arXiv:2606.09813},
  year={2026}
}

@article{zheng2026omnivta,
  title={{OmniVTA}: Visuo-Tactile World Modeling for Contact-Rich Robotic Manipulation},
  author={Zheng, Yuhang and Gu, Songen and Zheng, Yupeng and Li, Weize and Zang, Yujie and Tian, Shuai and Li, Xiang and Hao, Ce and Gao, Chen and Liu, Si and Li, Haoran and Chen, Yilun and Yan, Shuicheng and Ding, Wenchao},
  journal={arXiv preprint arXiv:2603.19201},
  year={2026}
}

@article{ma2026feelworld,
  title={{FeelWorld}: Visuo-Tactile World Model for Hierarchical Contact Prediction and Planning},
  author={Ma, Wenxuan and Zhang, Chaofan and Xue, Chao and Cai, Yinghao and Yao, Guocai and Cui, Shaowei and Wang, Shuo},
  journal={arXiv preprint arXiv:2607.24267},
  year={2026}
}

@article{wan2025,
  title={{Wan}: Open and Advanced Large-Scale Video Generative Models},
  author={{Wan Team} and Wang, Ang and Ai, Baole and Wen, Bin and Mao, Chaojie and Xie, Chen-Wei and Chen, Di and Yu, Feiwu and Zhao, Haiming and Yang, Jianxiao and others},
  journal={arXiv preprint arXiv:2503.20314},
  year={2025}
}

@inproceedings{rombach2022ldm,
  title={High-Resolution Image Synthesis with Latent Diffusion Models},
  author={Rombach, Robin and Blattmann, Andreas and Lorenz, Dominik and Esser, Patrick and Ommer, Bjorn},
  booktitle={Proceedings of the IEEE/CVF Conference on Computer Vision and Pattern Recognition},
  pages={10684--10695},
  year={2022}
}

@inproceedings{podell2024sdxl,
  title={{SDXL}: Improving Latent Diffusion Models for High-Resolution Image Synthesis},
  author={Podell, Dustin and English, Zion and Lacey, Kyle and Blattmann, Andreas and Dockhorn, Tim and M{\"u}ller, Jonas and Penna, Joe and Rombach, Robin},
  booktitle={International Conference on Learning Representations},
  year={2024}
}

@article{xu2025unit,
  title={{UniT}: Data Efficient Tactile Representation with Generalization to Unseen Objects},
  author={Xu, Zhengtong and Uppuluri, Raghava and Zhang, Xinwei and Fitch, Cael and Crandall, Philip Glen and Shou, Wan and Wang, Dongyi and She, Yu},
  journal={arXiv preprint arXiv:2408.06481},
  year={2025}
}

@article{zheng2024opensora,
  title={{Open-Sora}: Democratizing Efficient Video Production for All},
  author={Zheng, Zangwei and Peng, Xiangyu and Yang, Tianji and Shen, Chenhui and Li, Shenggui and Liu, Hongxin and Zhou, Yukun and Li, Tianyi and You, Yang},
  journal={arXiv preprint arXiv:2412.20404},
  year={2024}
}

@inproceedings{huang2024vbench,
  title={{VBench}: Comprehensive Benchmark Suite for Video Generative Models},
  author={Huang, Ziqi and He, Yinan and Yu, Jiashuo and Zhang, Fan and Si, Chenyang and Jiang, Yuming and Zhang, Yuanhan and Wu, Tianxing and Jin, Qingyang and Chanpaisit, Nattapol and Wang, Yaohui and Chen, Xinyuan and Wang, Limin and Lin, Dahua and Qiao, Yu and Liu, Ziwei},
  booktitle={Proceedings of the IEEE/CVF Conference on Computer Vision and Pattern Recognition},
  pages={21807--21818},
  year={2024}
}

@article{shao2024deepseekmath,
  title={{DeepSeekMath}: Pushing the Limits of Mathematical Reasoning in Open Language Models},
  author={Shao, Zhihong and Wang, Peiyi and Zhu, Qihao and Xu, Runxin and Song, Junxiao and Bi, Xiao and Zhang, Haowei and Zhang, Mingchuan and Li, Y. K. and Wu, Y. and Guo, Daya},
  journal={arXiv preprint arXiv:2402.03300},
  year={2024}
}

@inproceedings{zhao2023act,
  title={Learning Fine-Grained Bimanual Manipulation with Low-Cost Hardware},
  author={Zhao, Tony Z. and Kumar, Vikash and Levine, Sergey and Finn, Chelsea},
  booktitle={Robotics: Science and Systems},
  year={2023}
}

@inproceedings{lipman2023flowmatching,
  title={Flow Matching for Generative Modeling},
  author={Lipman, Yaron and Chen, Ricky T. Q. and Ben-Hamu, Heli and Nickel, Maximilian and Le, Matt},
  booktitle={International Conference on Learning Representations},
  year={2023}
}

@inproceedings{black2025pi05,
  title={{$\pi_{0.5}$}: A Vision-Language-Action Model with Open-World Generalization},
  author={Black, Kevin and Brown, Noah and Darpinian, James and Dhabalia, Karan and Driess, Danny and Esmail, Adnan and Equi, Michael Robert and Finn, Chelsea and Fusai, Niccolo and Galliker, Manuel Y. and Ghosh, Dibya and Groom, Lachy and Hausman, Karol and ichter, brian and Jakubczak, Szymon and Jones, Tim and Ke, Liyiming and LeBlanc, Devin and Levine, Sergey and Li-Bell, Adrian and Mothukuri, Mohith and Nair, Suraj and Pertsch, Karl and Ren, Allen Z. and Shi, Lucy Xiaoyang and Smith, Laura and Springenberg, Jost Tobias and Stachowicz, Kyle and Tanner, James and Vuong, Quan and Walke, Homer and Walling, Anna and Wang, Haohuan and Yu, Lili and Zhilinsky, Ury},
  booktitle={Proceedings of The 9th Conference on Robot Learning},
  series={Proceedings of Machine Learning Research},
  volume={305},
  pages={17--40},
  publisher={PMLR},
  year={2025},
  url={https://proceedings.mlr.press/v305/black25a.html}
}

@inproceedings{wang2025alignavideo,
  title={{Align-A-Video}: Deterministic Reward Tuning of Image Diffusion Models for Consistent Video Editing},
  author={Wang, Shengzhi and Zhong, Yingkang and Mu, Jiangchuan and Wu, Kai and Xiong, Mingliang and Fang, Wen and Liu, Mingqing and Deng, Hao and He, Bin and Li, Gang and Liu, Qingwen},
  booktitle={Proceedings of the IEEE/CVF Conference on Computer Vision and Pattern Recognition},
  pages={2074--2083},
  year={2025}
}

@article{yuan2026dextacwam,
  title={{DexTacWAM}: A Visuo-Tactile World-Action Model for Dexterous Manipulation},
  author={Yuan, Haoran and Wang, Zekai and Shao, Boning and Lu, Haoran and Darrell, Trevor and Lourentzou, Ismini and Zhan, Wei},
  journal={arXiv preprint arXiv:2609.24976},
  year={2026},
  url={https://arxiv.org/abs/2609.24976}
}

@article{liu2026taco,
  title={{TACO}: {TActile} World Model as a {Self-COrrector} for Scalable Robot Policy Post-Training},
  author={Liu, Shengbang and Jia, Yueru and Yan, Yuyang and Liu, Jiaming and Zhang, Xinran and Feng, Qiuxuan and Guo, Yandong and Zhou, Shiji and Shi, Boxin and Zhang, Shanghang},
  journal={arXiv preprint arXiv:2607.02840},
  year={2026},
  url={https://arxiv.org/abs/2607.02840}
}

@article{hafner2025dreamerv3,
  title={Mastering Diverse Control Tasks through World Models},
  author={Hafner, Danijar and Pasukonis, Jurgis and Ba, Jimmy and Lillicrap, Timothy},
  journal={Nature},
  volume={640},
  number={8059},
  pages={647--653},
  year={2025},
  doi={10.1038/s41586-025-08744-2},
  url={https://www.nature.com/articles/s41586-025-08744-2}
}

@article{wang2026interactive,
  title={Interactive World Simulator for Robot Policy Training and Evaluation},
  author={Wang, Yixuan and Syed, Rhythm and Wu, Fangyu and Zhang, Mengchao and Onol, Aykut and Barreiros, Jose and Nayyeri, Hooshang and Dear, Tony and Zhang, Huan and Li, Yunzhu},
  journal={arXiv preprint arXiv:2603.08546},
  year={2026},
  url={https://arxiv.org/abs/2603.08546}
}

@inproceedings{yang2026rise,
  title={Self-Improving Robot Policy with Compositional World Model},
  author={Yang, Jiazhi and Lin, Kunyang and Zhang, Wencong and Li, Jinwei and Lin, Tianwei and Wu, Longyan and Zhang, Ya-Qin and Zhao, Hao and Luo, Ping and Su, Zhizhong and Li, Hongyang and Yue, Xiangyu and Chen, Li},
  booktitle={Proceedings of Robotics: Science and Systems},
  year={2026},
  doi={10.15607/RSS.2026.XXII.012},
  url={https://www.roboticsproceedings.org/rss22/p012.html}
}

@article{yang2026mirabench,
  title={{MiraBench}: Evaluating Action-Conditioned Reliability in Robotic World Models},
  author={Yang, Tianzhuo and Shen, Zihan and Mi, Zirui and Zhang, Zhaoyi and Zhou, Jiayi and Ji, Jiaming and Dai, Juntao and Chen, Jiawei and Chen, Boyuan and Yang, Yaodong},
  journal={arXiv preprint arXiv:2605.29360},
  year={2026},
  url={https://arxiv.org/abs/2605.29360}
}

@inproceedings{bardhan2026persistent,
  title={Persistent Robot World Models: Stabilizing Multi-Step Rollouts via Reinforcement Learning},
  author={Bardhan, Jai and Drozdik, Patrik and Sivic, Josef and Petrik, Vladimir},
  booktitle={European Conference on Computer Vision},
  year={2026},
  url={https://arxiv.org/abs/2603.25685}
}

@inproceedings{rlinf,
  title={{RLinf}: Flexible and Efficient {Large-Scale} Reinforcement Learning via {Macro-to-Micro} Flow Transformation},
  author={Yu, Chao and Wang, Yuanqing and Guo, Zhen and Lin, Hao and Xu, Si and Zang, Hongzhi and Zhang, Quanlu and Wu, Yongji and Zhu, Chunyang and Hu, Junhao and Huang, Zixiao and Wei, Mingjie and Xie, Yuqing and Yang, Ke and Dai, Bo and Xu, Zhexuan and Du, Jiakun and Wang, Xiangyuan and Fu, Xu and Shi, Letong and Liu, Zhihao and Chen, Kang and Liu, Weilin and Liu, Gang and Li, Boxun and Yang, Jianlei and Yang, Zhi and Dai, Guohao and Wang, Yu},
  booktitle={20th USENIX Symposium on Operating Systems Design and Implementation (OSDI 26)},
  pages={829--846},
  year={2026},
  publisher={USENIX Association},
  url={https://www.usenix.org/conference/osdi26/presentation/yu-chao}
}

@inproceedings{ha2018worldmodels,
  title={Recurrent World Models Facilitate Policy Evolution},
  author={Ha, David and Schmidhuber, J{\"u}rgen},
  booktitle={Advances in Neural Information Processing Systems},
  volume={31},
  year={2018},
  url={https://papers.nips.cc/paper/2018/hash/2de5d16682c3c35007e4e92982f1a2ba-Abstract.html}
}

@inproceedings{hafner2020dreamer,
  title={Dream to Control: Learning Behaviors by Latent Imagination},
  author={Hafner, Danijar and Lillicrap, Timothy and Ba, Jimmy and Norouzi, Mohammad},
  booktitle={International Conference on Learning Representations},
  year={2020},
  url={https://arxiv.org/abs/1912.01603}
}

@inproceedings{hansen2024tdmpc2,
  title={{TD-MPC2}: Scalable, Robust World Models for Continuous Control},
  author={Hansen, Nick and Su, Hao and Wang, Xiaolong},
  booktitle={International Conference on Learning Representations},
  year={2024},
  url={https://proceedings.iclr.cc/paper_files/paper/2024/hash/cf73d57b6dcda32b293df7c2d5341f49-Abstract-Conference.html}
}

@inproceedings{wu2023daydreamer,
  title={{DayDreamer}: World Models for Physical Robot Learning},
  author={Wu, Philipp and Escontrela, Alejandro and Hafner, Danijar and Abbeel, Pieter and Goldberg, Ken},
  booktitle={Proceedings of The 6th Conference on Robot Learning},
  series={Proceedings of Machine Learning Research},
  volume={205},
  pages={2226--2240},
  publisher={PMLR},
  year={2023},
  url={https://proceedings.mlr.press/v205/wu23c.html}
}

@inproceedings{micheli2023iris,
  title={Transformers are Sample-Efficient World Models},
  author={Micheli, Vincent and Alonso, Eloi and Fleuret, Fran{\c{c}}ois},
  booktitle={International Conference on Learning Representations},
  year={2023},
  url={https://arxiv.org/abs/2209.00588}
}

@inproceedings{bruce2024genie,
  title={Genie: Generative Interactive Environments},
  author={Bruce, Jake and Dennis, Michael D and Edwards, Ashley and Parker-Holder, Jack and Shi, Yuge and Hughes, Edward and Lai, Matthew and Mavalankar, Aditi and Steigerwald, Richie and Apps, Chris and Aytar, Yusuf and Bechtle, Sarah Maria Elisabeth and Behbahani, Feryal and Chan, Stephanie C.Y. and Heess, Nicolas and Gonzalez, Lucy and Osindero, Simon and Ozair, Sherjil and Reed, Scott and Zhang, Jingwei and Zolna, Konrad and Clune, Jeff and Freitas, Nando De and Singh, Satinder and Rockt{\"a}schel, Tim},
  booktitle={Proceedings of the 41st International Conference on Machine Learning},
  series={Proceedings of Machine Learning Research},
  volume={235},
  pages={4603--4623},
  publisher={PMLR},
  year={2024},
  url={https://proceedings.mlr.press/v235/bruce24a.html}
}

@inproceedings{ho2022videodiffusion,
  title={Video Diffusion Models},
  author={Ho, Jonathan and Salimans, Tim and Gritsenko, Alexey and Chan, William and Norouzi, Mohammad and Fleet, David J.},
  booktitle={Advances in Neural Information Processing Systems},
  year={2022},
  url={https://arxiv.org/abs/2204.03458}
}

@inproceedings{blattmann2023alignlatents,
  title={Align Your Latents: High-Resolution Video Synthesis with Latent Diffusion Models},
  author={Blattmann, Andreas and Rombach, Robin and Ling, Huan and Dockhorn, Tim and Kim, Seung Wook and Fidler, Sanja and Kreis, Karsten},
  booktitle={Proceedings of the IEEE/CVF Conference on Computer Vision and Pattern Recognition},
  year={2023},
  url={https://arxiv.org/abs/2304.08818}
}

@article{lambeta2020digit,
  title={{DIGIT}: A Novel Design for a Low-Cost Compact High-Resolution Tactile Sensor with Application to In-Hand Manipulation},
  author={Lambeta, Mike and Chou, Po-Wei and Tian, Stephen and Yang, Brian and Maloon, Benjamin and Most, Victoria Rose and Stroud, Dave and Santos, Raymond and Byagowi, Ahmad and Kammerer, Gregg and Jayaraman, Dinesh and Calandra, Roberto},
  journal={IEEE Robotics and Automation Letters},
  volume={5},
  number={3},
  pages={3838--3845},
  year={2020},
  url={https://arxiv.org/abs/2005.14679}
}

@article{yuan2017gelsight,
  title={{GelSight}: High-Resolution Robot Tactile Sensors for Estimating Geometry and Force},
  author={Yuan, Wenzhen and Dong, Siyuan and Adelson, Edward H.},
  journal={Sensors},
  volume={17},
  number={12},
  pages={2762},
  year={2017},
  doi={10.3390/s17122762}
}

@article{lee2020making,
  title={Making Sense of Vision and Touch: Learning Multimodal Representations for Contact-Rich Tasks},
  author={Lee, Michelle A. and Zhu, Yuke and Zachares, Peter and Tan, Matthew and Srinivasan, Krishnan and Savarese, Silvio and Fei-Fei, Li and Garg, Animesh and Bohg, Jeannette},
  journal={IEEE Transactions on Robotics},
  volume={36},
  number={3},
  year={2020},
  url={https://arxiv.org/abs/1907.13098}
}

@inproceedings{guzey2024seetotouch,
  title={See to Touch: Learning Tactile Dexterity through Visual Incentives},
  author={Guzey, Irmak and Dai, Yinlong and Evans, Ben and Chintala, Soumith and Pinto, Lerrel},
  booktitle={2024 IEEE International Conference on Robotics and Automation},
  pages={13825--13832},
  year={2024},
  doi={10.1109/ICRA57147.2024.10611407},
  url={https://ieeexplore.ieee.org/document/10611407/}
}

@inproceedings{peebles2023dit,
  title={Scalable Diffusion Models with Transformers},
  author={Peebles, William and Xie, Saining},
  booktitle={Proceedings of the IEEE/CVF International Conference on Computer Vision},
  pages={4195--4205},
  year={2023},
  url={https://arxiv.org/abs/2212.09748}
}

@inproceedings{he2016resnet,
  title={Deep Residual Learning for Image Recognition},
  author={He, Kaiming and Zhang, Xiangyu and Ren, Shaoqing and Sun, Jian},
  booktitle={Proceedings of the IEEE Conference on Computer Vision and Pattern Recognition},
  pages={770--778},
  year={2016},
  doi={10.1109/CVPR.2016.90}
}
\bibliographystyle{iclr2027_conference}
\endgroup

\clearpage
\appendix
\section*{Appendix}
\label{appendix}

\startcontents[app]
\begingroup
  \renewcommand{\contentsname}{Appendix Contents}
  \etocsettocdepth{subsection}
  \printcontents[app]{l}{0}{}
\endgroup
\newpage
\section{Hardware Configuration}
\label{app:hardware}
\noindent
Our experiments use an ARX AC One dual-arm platform with a tabletop workspace. Each arm has six joints and a gripper. Three cameras provide a fixed base view and two wrist views, while four Xense optical tactile sensors mounted on the gripper fingers provide contact observations. Visual, tactile, and robot-state observations are synchronized at 10\,Hz.

\begin{center}
\includegraphics[width=.94\linewidth]{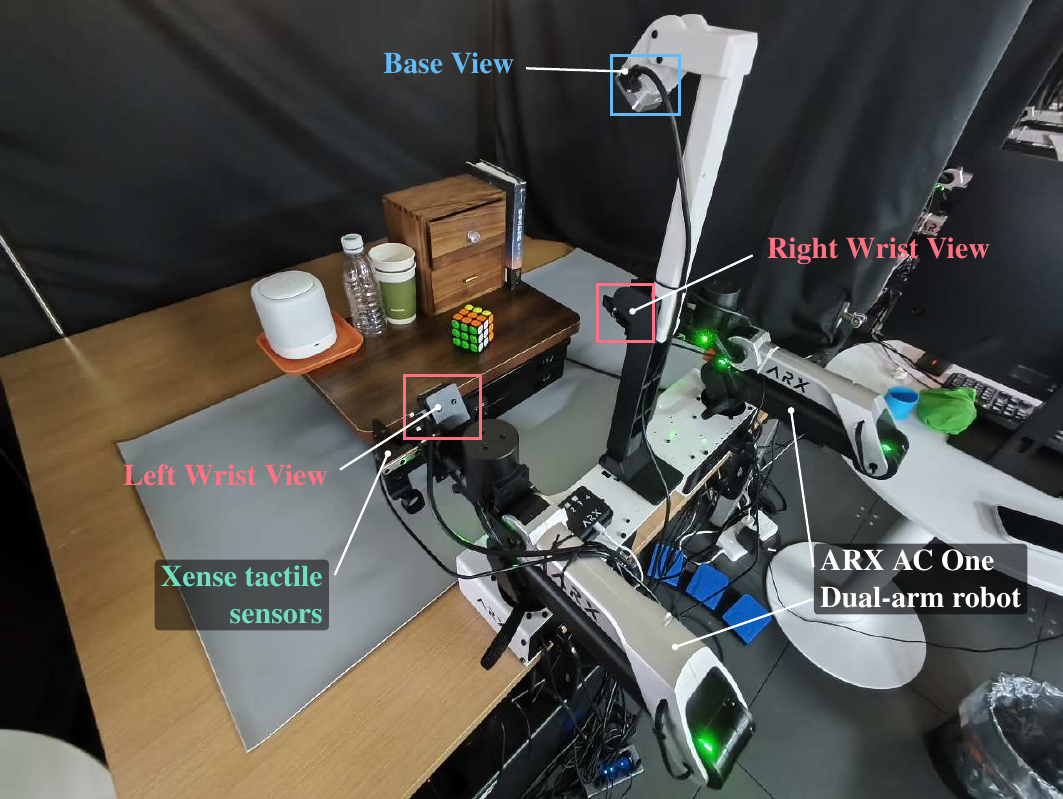}
\captionof{figure}{\textbf{Hardware layout.} The dual-arm platform, three camera views, and fingertip tactile sensors.}
\label{fig:hardware_layout}
\end{center}

\noindent\textbf{Optical tactile sensing.}
Xense sensors capture local contact appearance and surface deformation at the fingertips. We use tactile images and marker-displacement fields as complementary observations of the interaction. The photographs below show the sensor's contact-facing side, rear connector and mounting holes, and installation on a gripper finger.

\begin{center}
\begin{minipage}[t]{.31\linewidth}
\centering
\includegraphics[width=\linewidth,height=110pt,keepaspectratio]{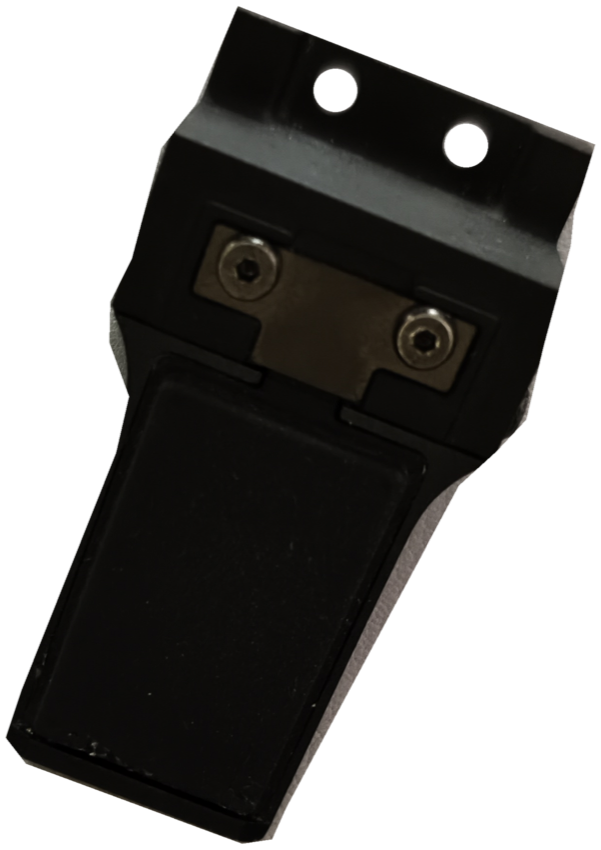}
\par\smallskip{\small (a) Contact-facing side}
\end{minipage}\hfill
\begin{minipage}[t]{.31\linewidth}
\centering
\includegraphics[width=\linewidth,height=110pt,keepaspectratio]{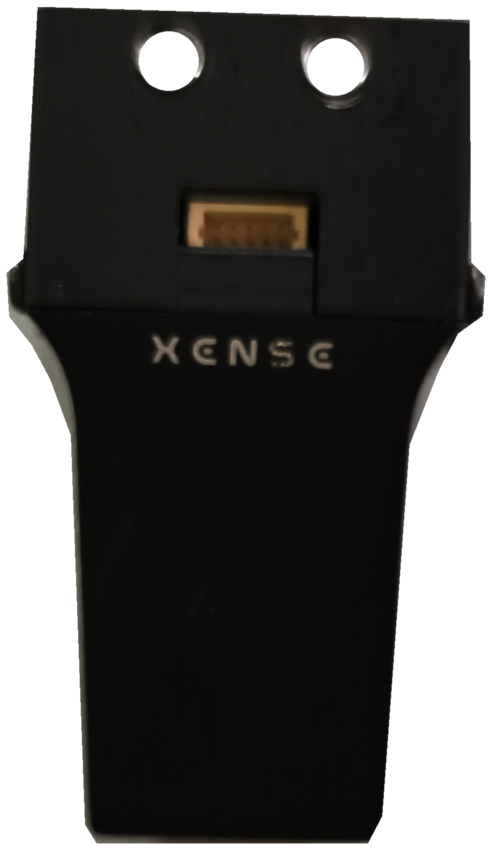}
\par\smallskip{\small (b) Rear and connector}
\end{minipage}\hfill
\begin{minipage}[t]{.34\linewidth}
\centering
\includegraphics[width=\linewidth,height=110pt,keepaspectratio]{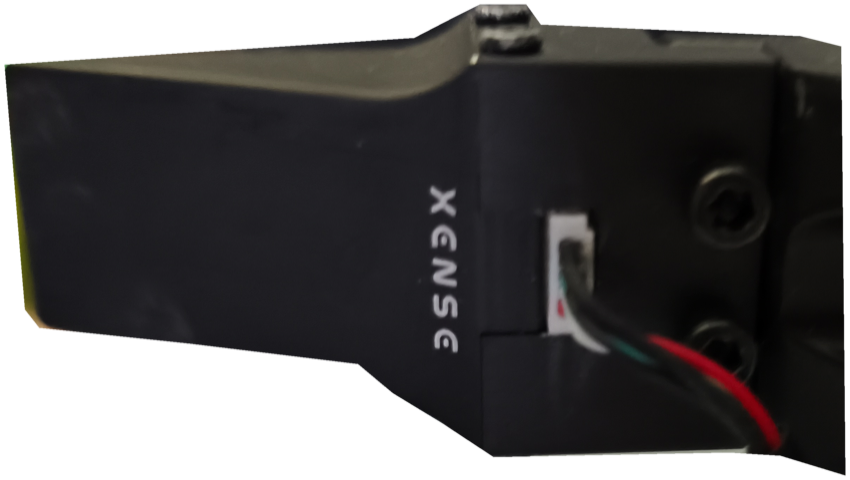}
\par\smallskip{\small (c) Gripper installation}
\end{minipage}
\captionof{figure}{\textbf{Xense optical tactile sensors.} Sensor views and gripper mounting.}
\label{fig:hardware_tactile}
\end{center}

\FloatBarrier
\section{Bimanual Tasks and Data}
\label{app:tasks_data}
\subsection{Task design and object selection}
All four tasks require coordinated use of both arms. We select objects that expose complementary contact phenomena: reorientation of a slender object, separation of interlocking parts, threaded rotation, and changing surface textures. Figure~\ref{fig:task_raw_cases} pairs object photographs with robot rollouts from our data and synchronized tactile observations.

\paragraph{Test Tube.}
The left gripper picks up a test tube and passes it to the right gripper. The right gripper then uses contact with the table to reorient the tube in-hand before inserting it into the rack. The slender, transparent tube makes grasp stability and orientation difficult to infer from appearance alone, motivating tactile feedback during handover and reorientation.

\paragraph{Blocks.}
The left gripper lifts two connected blocks. Both grippers separate them and place each block into the bag of the matching color. Interlocking studs and edges create localized contact patterns, while separation produces a distinct change in contact that must agree with the visible motion.

\paragraph{Bottle.}
One gripper stabilizes the bottle while the other unscrews its cap; the cap and bottle are then placed separately on the table. The ridged cap and compliant bottle provide different contact geometries. This task couples rotational motion with sustained grasping and a final contact-release event.

\paragraph{Rubik's Cube.}
The robot restores a Rubik's Cube after a single face-turn perturbation. Its colored, textured faces change the local contact pattern during rotation, making it a challenging test of visual--tactile alignment. The restricted scramble isolates coordinated manipulation and sensory consistency rather than general Rubik's Cube solution planning.

\begin{center}
\includegraphics[width=\linewidth]{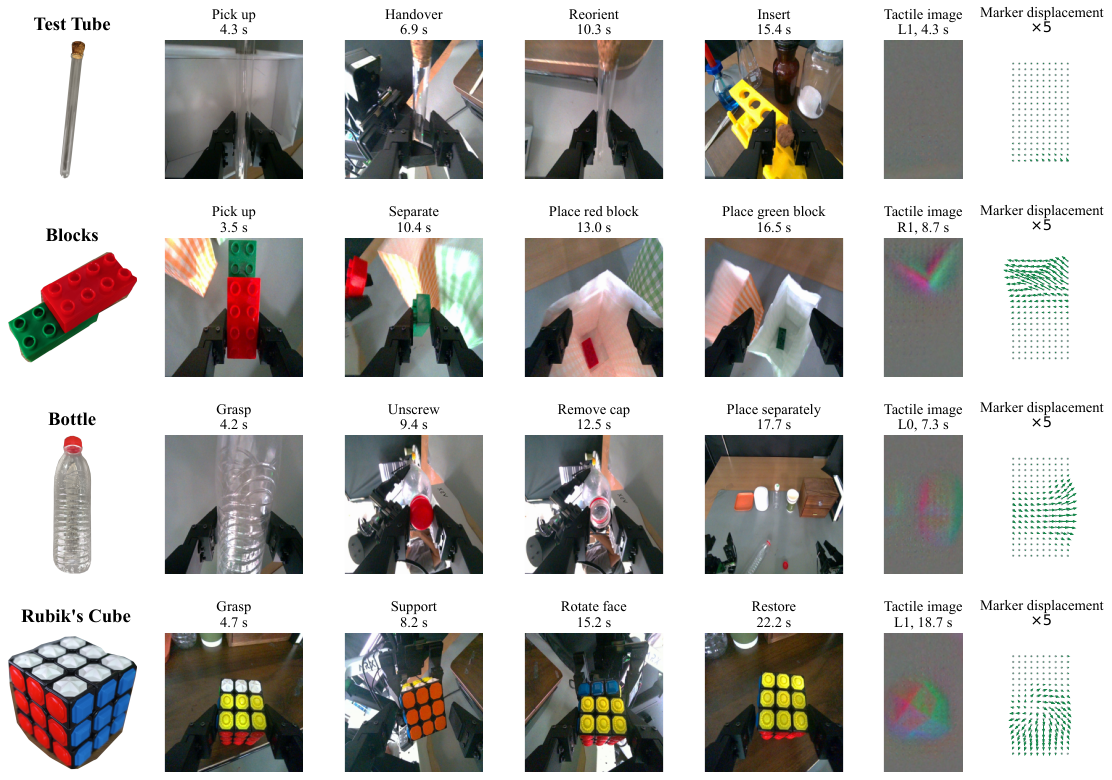}
\captionof{figure}{\textbf{Four bimanual tasks.} Object photographs, rollout stages, and paired tactile-image/marker-displacement examples. L/R denotes the gripper and 0/1 its pad; green arrows show recorded displacement at $5\times$ display scale.}
\label{fig:task_raw_cases}
\end{center}

\clearpage
\subsection{Data scale and tactile representation}
\label{app:data_scale}
\paragraph{Data.}
We use 3,805 real-robot policy rollouts for world-model training and 80 held-out test rollouts per task (Table~\ref{tab:task_data_scale}).

\begin{center}
\captionof{table}{\textbf{Policy-rollout training and test data used by the world model.}}
\label{tab:task_data_scale}
\small
\begin{tabular}{lrr}
\toprule
Task & Training episodes & Test episodes \\
\midrule
Test Tube & 1,000 & 80 \\
Blocks & 1,000 & 80 \\
Bottle & 926 & 80 \\
Rubik's Cube & 879 & 80 \\
\midrule
Total & 3,805 & 320 \\
\bottomrule
\end{tabular}
\end{center}

\paragraph{Visual observation.}
Each timestamp contains three synchronized RGB views: one fixed base view and two wrist views. Observations are recorded at 10\,Hz. Each view is cropped and resized to $224\times224$ pixels, then the three views are concatenated horizontally in a fixed order to form a $224\times672$ visual strip.

\begin{center}
\includegraphics[width=\linewidth]{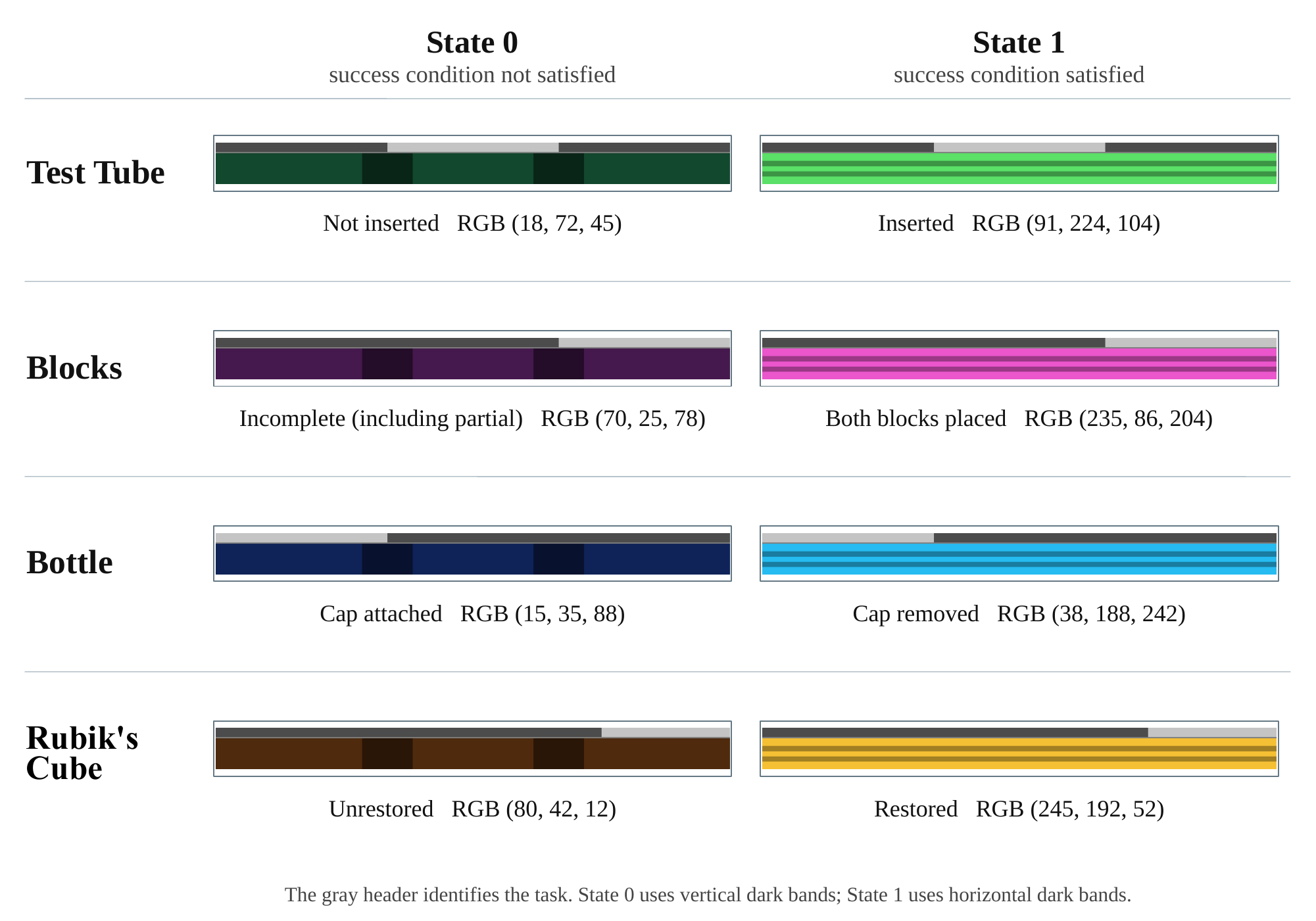}
\captionof{figure}{\textbf{Task-state canvas encoding.} Each row shows state~0 (the success condition is not satisfied) and state~1 (the success condition is satisfied) for one task. Task-specific color identifies the task, while vertical and horizontal low-frequency bands distinguish states~0 and~1, respectively. For Blocks, partial completion remains state~0; state~1 requires both blocks to be placed.}
\label{fig:task_state_canvas}
\end{center}

\paragraph{Tactile image.}
We use the Xense internal camera's recorded difference image relative to a sensor reference, not consecutive-frame differences or external scene RGB. Preprocessing decodes and time-aligns these $400\times700\times3$ unsigned 8-bit images. Display rotates each pad $90^\circ$ clockwise; the world model does not re-estimate the SDK's reference or intensity mapping.

\paragraph{Marker displacement.}
Each pad stores a $35\times20$ grid of 700 tracked markers. For marker $i$, the first frame of an episode defines the reference state $\mathbf q_{0,i}$. At time $t$, we subtract this reference from the recorded three-component marker state $\mathbf q_{t,i}$ to obtain $\Delta\mathbf q_{t,i}=(\Delta x,\Delta y,\Delta z)$. A stationary marker is therefore centered at zero, whereas contact-induced translation or deformation produces a signed three-dimensional offset. To place this nonvisual field on the shared RGB canvas, we clip each component to $[-s,s]$, normalize it to $[0,1]$, and assign $x$, $y$, and $z$ to the R, G, and B channels, respectively:
\begin{equation}
\Delta\mathbf q_{t,i}=\mathbf q_{t,i}-\mathbf q_{0,i},\qquad
\mathbf c_{t,i}=\tfrac12\left[\operatorname{clip}(\Delta\mathbf q_{t,i}/s,-1,1)+1\right].
\label{eq:appendix_flow_encoding}
\end{equation}
The audited configurations use $s=0.06$ in cache-native units; this is an encoding scale rather than an observed physical bound. The RGB image is only an image-native representation for the model: quantitative statistics are computed directly from the unencoded $\Delta\mathbf q$. For vector visualization, the initial planar marker locations provide the arrow origins and $(\Delta x,\Delta y)$ gives the sensor-plane arrow direction; $\Delta z$ remains part of the three-dimensional field and its RGB encoding. Only the displayed grid is subsampled to $18\times10$, while statistics retain all 700 markers.

\noindent\begin{minipage}[t]{.58\linewidth}
\textbf{Action representation.}
The action condition is a numerical sequence of arm-joint positions and gripper states. At each timestamp, we apply it to the robot URDF, recover the articulated meshes through forward kinematics, and project them into the fixed and wrist cameras using calibrated intrinsics and extrinsics. Each render uses the same crop, resize, and view order as RGB, producing a spatially aligned action-image sequence. The shared video VAE encodes these images before position-wise fusion with the noisy world-model latents. The overlay at right visualizes registration and is not a prediction target.
\end{minipage}\hfill
\begin{minipage}[t]{.38\linewidth}
\vspace{0pt}\centering
\includegraphics[width=\linewidth]{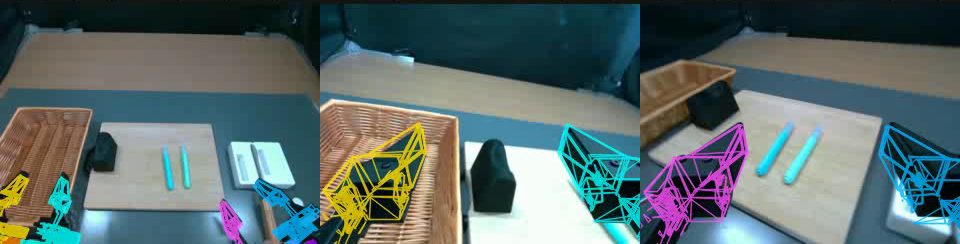}
{\footnotesize\captionof{figure}{\textbf{Action projection.} Projected URDF wireframes over the fixed and wrist views.}\label{fig:action_projection_appendix}}
\end{minipage}

\paragraph{Tactile ranges.}
We summarize tactile-image intensities and marker-displacement magnitudes from sampled rollouts across all four tasks. The P95 displacement norms are 0.0043, 0.0161, 0.0234, and 0.0266 for Test Tube, Blocks, Bottle, and Rubik's Cube, respectively, in cache-native units (Figure~\ref{fig:tactile_raw_statistics}).

\begin{center}
\includegraphics[width=.96\linewidth]{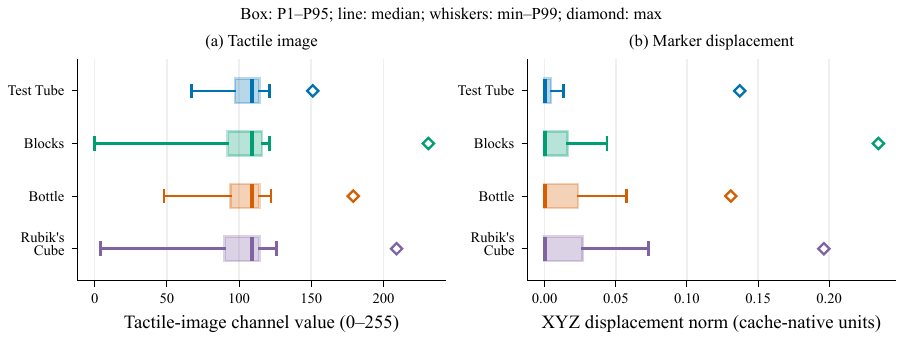}
\captionof{figure}{\textbf{Tactile data distributions.} Unified percentile-box summaries from sampled rollouts.}
\label{fig:tactile_raw_statistics}
\end{center}

\section{World Model Training and Inference}
\label{app:world_model}

\subsection{World-model data collection}
\label{app:wm_data_collection}
To expose the world model to a broad action distribution, we execute stochastic rollouts of task policies on the real robot.  The resulting data contain successful executions as well as naturally occurring failures, including missed grasps, premature release, inaccurate handover, misalignment, and incomplete insertion or placement.  This diversity is important because a policy evaluator must model not only expert demonstrations but also the off-nominal states visited by imperfect policies.

\begin{figure}[h]
\centering
\includegraphics[width=\linewidth]{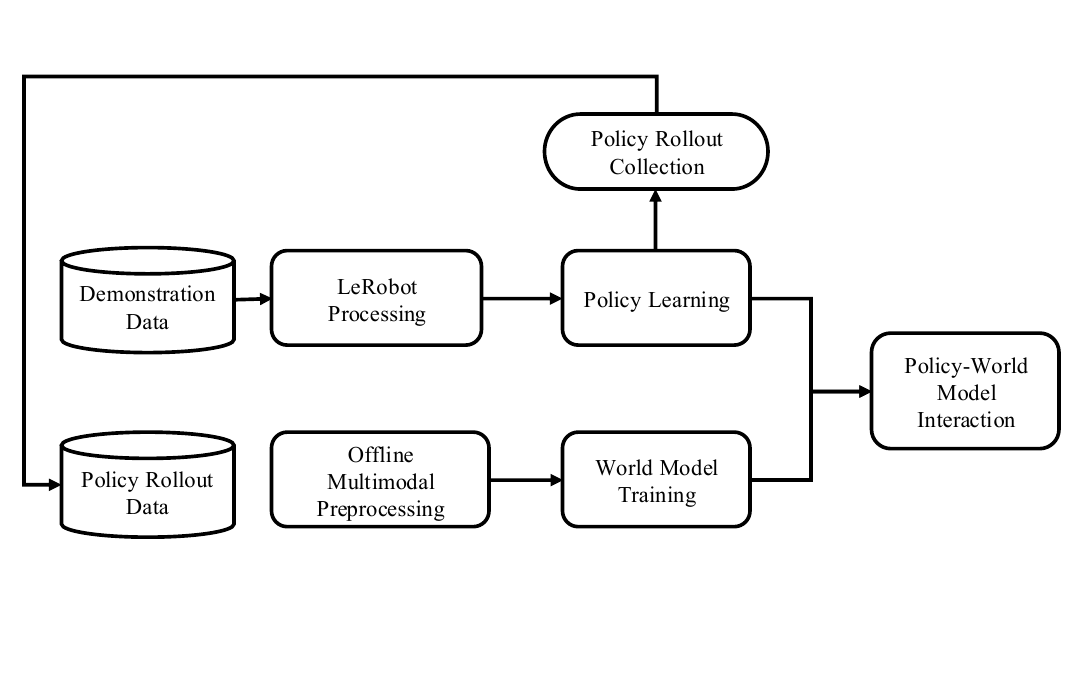}
\caption{\textbf{World-model data and interaction loop.} Policy rollouts are processed offline into aligned action-image and visual--tactile training data.  The trained world model then interacts with a policy for RL or evaluation.}
\label{fig:wm_data_pipeline}
\end{figure}

Figure~\ref{fig:wm_data_pipeline} summarizes the data loop.  Robot demonstrations are first converted to the LeRobot format and used to train task policies.  Stochastic policy execution then provides rollouts for world-model learning.  After collection, these rollouts undergo offline multimodal preprocessing: (1) numerical robot actions are rendered as view-aligned action images; (2) camera images are cropped and resized to the fixed world-model layout; and (3) each raw tactile record is separated into its tactile image and tracked marker positions, from which marker displacement is computed.  We time-align these processed signals and assemble them into the unified canvas before world-model training.  The trained world model subsequently interacts with the policy in closed loop, providing generated visual--tactile observations for either policy evaluation or RL-based policy improvement.

\subsection{Training configuration}
\label{app:wm_training}
\paragraph{Initialization and action conditioning.}
We initialize \method{} from \texttt{Wan2.2-TI2V-5B}, a 5-billion-parameter latent video flow-matching transformer pretrained for text/image-to-video generation.  We retain its visual VAE and spatiotemporal transformer, freeze the VAE, and fine-tune the transformer for multimodal future prediction.  As described in Section~\ref{sec:conditioning}, robot configurations are rendered into view-aligned action images using camera calibration and the robot URDF.  Their latent tokens are fused position-wise with the noisy canvas tokens, converting the pretrained generator into an action-conditioned world model without a dedicated action encoder beyond the shared video VAE.

\paragraph{Optimization and compute.}
Training uses 64 PPU-ZW810E accelerators (four nodes with 16 devices each) for 300 epochs and takes approximately three days.  We use the 8-bit CAME optimizer with a constant learning rate of $2^{-14.5}=4.316\times10^{-5}$ and weight decay $10^{-2}$.  The per-device batch size is 2 with one gradient-accumulation step, giving an effective batch size of 128 clips.  Training uses BF16, DeepSpeed ZeRO-2, and activation checkpointing of Wan transformer blocks.  No explicit gradient clipping is applied.

\paragraph{Noise and rollout training.}
For each flow-matching target, we sample $u\sim\mathcal U(0,1)$ and use the shifted noise level
\begin{equation}
  \sigma=\frac{5u}{1+4u},
  \qquad
  \mathbf z_{\sigma}=(1-\sigma)\mathbf z+\sigma\boldsymbol\epsilon,
  \qquad \boldsymbol\epsilon\sim\mathcal N(0,I),
  \label{eq:appendix_wm_noise}
\end{equation}
where $\sigma$ is quantized to increments of $10^{-3}$. Each training example retains the episode-initial three-view RGB anchor across four successive rollout chunks. A chunk predicts eight future frames; its terminal observation replaces only the current multimodal reference $\mathbf r$ for the next chunk. The loss is averaged across the four chunks, so the model is optimized under its own recursively generated context rather than only under ground-truth one-step conditioning.

\subsection{Inference and recursive rollout}
\label{app:wm_inference}
At inference, the same episode-initial RGB anchor is fixed across all chunks. Each chunk starts from Gaussian noise at $\sigma=1$ and is denoised with 5 flow-matching steps to generate eight future multimodal frames; its last frame updates $\mathbf r$ for the next call. In interactive evaluation, the policy consumes the generated RGB and tactile observations and supplies the next action sequence; the action is rendered into the aligned conditioning canvas before the following chunk is generated. We repeat this process until the full episode horizon is reached, so rollout length is not limited to the four chunks used during training.

\FloatBarrier
\section{Policy Architectures and Tactile Integration}
\label{app:policy_architectures}

We evaluate three policy families: Action Chunking with Transformers (ACT)~\citep{zhao2023act}, a task-specific flow-matching policy (FM)~\citep{lipman2023flowmatching}, and a tactile-augmented $\pi_{0.5}$ policy~\citep{black2025pi05}. All three receive three synchronized RGB views, the current bimanual robot state, and four tactile pads, and output continuous dual-arm joint commands at 10\,Hz. Because tactile-to-policy fusion is not the primary focus of this work, we deliberately adopt a simple modality-concatenation strategy rather than designing a specialized fusion architecture. The three policies nevertheless differ in their pretraining, action parameterization, and data sources. FM and ACT are trained separately for each task from successful real-robot executions, whereas $\pi_{0.5}$ shares one multitask checkpoint across language instructions. We therefore use them as deployed policy families rather than as a controlled architecture comparison.

\subsection{Common observation and control interface}
\label{app:policy_common_interface}

The RGB observation contains one base-camera view and two wrist-camera views, each resized to $224\!\times\!224$. FM and ACT encode each view with a frozen DINOv2-small backbone and use its 384-dimensional CLS feature. The tactile input uses marker displacement rather than the tactile image: for each of four Xense pads, 700 tracked markers form a $35\!\times\!20$ grid with an XYZ displacement vector relative to the episode reference. FM and ACT concatenate the four pads along the channel dimension into a $35\!\times\!20\!\times\!12$ tensor; $\pi_{0.5}$ retains four separate $35\!\times\!20\!\times\!3$ grids. Each policy observes only the current synchronized frame and the current 14-dimensional dual-arm joint state. The robot interface consumes the first 14 output dimensions as absolute joint targets.

\begin{figure}[!t]
\centering
\includegraphics[width=\linewidth]{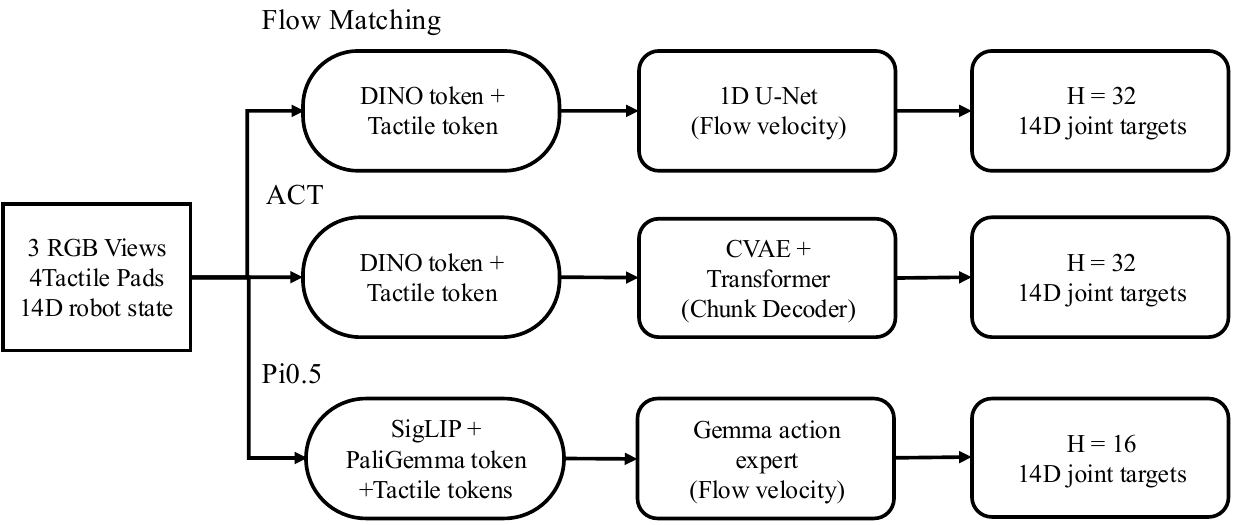}
\caption{\textbf{Policy architecture overview.} All policies use synchronized multi-view RGB, marker displacement, and robot state, but fuse these modalities differently. FM and ACT are task-specific; $\pi_{0.5}$ is prompt-conditioned and shared across tasks.}
\label{fig:policy_architecture_overview}
\end{figure}

Figure~\ref{fig:policy_architecture_overview} summarizes how the same sensory interface is specialized by the three policies. FM compresses vision, touch, and state into one global condition for a one-dimensional U-Net velocity field. ACT exposes the modality features as Transformer memory tokens and decodes a complete action chunk with learned queries. $\pi_{0.5}$ appends four tactile tokens to its vision--language prefix and predicts actions with a Gemma action expert.

\subsection{Flow-matching policy}
\label{app:policy_fm}

The FM policy encodes the three RGB views with frozen DINOv2-small features and uses lightweight encoders for the 14D robot state and the concatenated tactile-displacement grid. The three modalities are fused into a 256D condition \(c_t\). A conditional 1D U-Net then learns a flow field from Gaussian noise to a normalized relative \(32 \times 14\) joint-action chunk using the standard flow-matching objective. At inference, we integrate the learned flow field with 32 Euler steps. The predicted chunk is unnormalized using the checkpoint action normalizer and converted, relative to the current robot state, into absolute 14D joint targets. All 32 actions are executed at 10 Hz before the policy replans.

\subsection{ACT}
\label{app:policy_act}

Our ACT variant uses frozen DINOv2-small features for the three RGB views and compresses the four tactile grids into a single token. These modality tokens, together with the current robot state and a CVAE latent, condition a Transformer decoder with learned action queries. The model is trained to reconstruct normalized relative joint-action chunks with KL regularization on the CVAE latent. At deployment, the latent is fixed to zero and the complete action chunk is decoded in one forward pass. The predicted relative actions are unnormalized using the checkpoint action normalizer and converted, relative to the current robot state, into absolute 14D joint targets before execution. For all four tasks, ACT predicts and executes a 32-step action chunk at 10 Hz before replanning, without temporal ensembling.

\subsection{Tactile-augmented $\pi_{0.5}$}
\label{app:policy_pi05}

The $\pi_{0.5}$ policy combines SigLIP visual tokens, a PaliGemma language--state context branch, and a Gemma action expert for flow-based generation. A shared CNN independently encodes each tactile pad into a token, and the four tactile tokens are appended to the language--vision prefix through a learned gate. The model predicts a 16-step action chunk in a 32D internal space, of which the first 14 dimensions control the robot. Deterministic evaluation and real-robot deployment use a ten-step Flow-ODE sampler from Gaussian noise to action. During GRPO rollout, one of the ten denoising steps is selected uniformly and replaced by a stochastic Flow-SDE transition, while the remaining steps use Flow-ODE integration.

\FloatBarrier
\section{Reward Model Training}
\label{app:reward_models}

\subsection{Roles and Supervision}
\label{app:reward_roles}

We train reward models to provide supervision for world-model post-training and rewards for policy reinforcement learning inside the trained world model.  Three models address complementary aspects of the generated interaction.  The State Reward Model scores whether an RGB observation satisfies the task outcome; its scores guide world-model post-training and provide the binary reward for policy GRPO.  The Scene Reward Model scores whether a generated frame preserves the initial scene while allowing intended robot and object motion.  The Modality Reward Model scores whether visual and tactile observations describe a compatible physical state.  The latter two provide scene and cross-modal supervision during world-model post-training.

All three models output a binary logit and are optimized with \texttt{BCEWithLogitsLoss}; a sigmoid is applied only when a score in $[0,1]$ is needed.  Table~\ref{tab:reward_model_data} summarizes their supervision.  The task-state labels come entirely from real robot rollouts: human annotators identify the transition into a successful state, frames before that transition are negative, and unsuccessful rollouts remain negative throughout.  Scene-consistency supervision combines a real initial-frame reference with world-model-generated rollouts; annotators mark the first generated frame that no longer agrees with the reference scene.  Visual--tactile supervision starts from synchronized real multimodal bundles: aligned observations are positive, while within-hand tactile swaps and tactile bundles drawn from another rollout create controlled negative examples.  We also add world-model-generated observations judged to have mismatched visual and tactile content as negative examples.  These sources expose the model to both constructed modality conflicts and generation failures.

\begin{table}[H]
\centering
\small
\caption{\textbf{Reward-model supervision.} ``WM'' denotes world-model-generated observations and ``GT'' denotes real observations.}
\label{tab:reward_model_data}
\begin{tabularx}{\linewidth}{@{}p{0.20\linewidth}p{0.35\linewidth}X@{}}
\toprule
\textbf{Reward model} & \textbf{Input and question} & \textbf{Data provenance} \\
\midrule
State Reward Model & Three RGB views; is the task complete? & Human-labeled real rollouts with frame-level completion transitions. \\
Scene Reward Model & Initial GT and current WM views; is the scene preserved? & Human-labeled GT--WM pairs, including preserved and drifted frames. \\
Modality Reward Model & RGB, tactile image, and marker displacement; do modalities agree? & Aligned real bundles, controlled tactile mismatches, and labeled WM generation failures. \\
\bottomrule
\end{tabularx}
\end{table}

\subsection{Network Architecture}
\label{app:reward_architecture}

Following the image-based reward-model design in RLinf~\cite{rlinf}, we use ImageNet-initialized ResNet-18 encoders and a scalar scoring head.  The State Reward Model shares one encoder across three RGB views and scores their concatenated features with a $1536\!\rightarrow\!256\!\rightarrow\!1$ MLP.  The Scene Reward Model shares an encoder between the initial GT and current WM views; a two-layer Transformer decoder compares their $7\times7$ spatial tokens, using the reference as memory.  The Modality Reward Model uses separate encoders for RGB, tactile image, and marker displacement, shared across views or pads within each modality.  Its four-layer decoder lets visual tokens query tactile tokens.  Both decoders use eight attention heads and a $512\!\rightarrow\!256\!\rightarrow\!1$ scoring head.

\subsection{Training Configuration}
\label{app:reward_training}

We train all three reward models for 100 epochs with AdamW on PPU or H20 accelerators.  The State Reward Model and Scene Reward Model use balanced positive and negative batches, whereas the Modality Reward Model retains the mixture of aligned observations, controlled tactile mismatches, and world-model-generated failures described above.  We use a learning rate of $10^{-4}$ for task state and scene consistency and $2\!\times\!10^{-4}$ for visual--tactile consistency.  Data are split by rollout to keep observations from the same trajectory together.  Once trained, the reward models are frozen during world-model post-training and policy GRPO.

\FloatBarrier
\section{Policy--World Model Interaction Rollout}
\label{app:policy_wm_rollout}

\subsection{Closed-loop interaction}
\label{app:policy_wm_closed_loop}

A policy rollout in \method{} treats the world model as an action-driven virtual environment.  A trajectory begins from one synchronized real observation, after which the policy predicts an action chunk from the current observation. We use a common task specification $\xi$: it identifies the task for task-specific FM and ACT, and includes the language instruction required by the shared $\pi_{0.5}$ policy. For all policy families, the corresponding task prompt is independently supplied to the text-conditioned world model. The world model executes each chunk in short segments and generates the corresponding future observations. Its latest prediction is converted back into the policy observation interface, closing the loop until the target horizon is reached. Except for the initial state, the policy therefore acts on generated rather than recorded future observations.

This interaction extends visual world-model rollout to the sensory interface required by contact-rich policies.  In addition to multi-view RGB, the generated canvas carries tactile images, marker displacement, and task-state information.  The policy consumes its required subset of these synchronized signals together with the current robot state.  Consequently, policy behavior can respond not only to visible geometry but also to predicted contact changes, while the same simulator provides a common interaction loop for policies with different action horizons.

\Needspace{26\baselineskip}
\subsection{Rollout configuration and execution}
\label{app:policy_wm_configuration}

\begin{wrapfigure}{r}{0.48\textwidth}
\vspace{-\intextsep}
\centering
\begin{minipage}{\linewidth}
\small
\refstepcounter{physcanvasalgorithm}
\label{alg:policy_wm_rollout}
\hrule height 0.6pt
\vspace{0.3ex}
\noindent\textbf{Algorithm~\thephyscanvasalgorithm\quad Closed-loop rollout}
\par\vspace{0.3ex}\hrule height 0.35pt
\vspace{0.5ex}
\renewcommand{\arraystretch}{1.08}
\begin{tabular}{@{}r@{\hspace{0.55em}}l@{}}
 1: & $Y\leftarrow[\,],\ q\leftarrow q_0,\ r\leftarrow o_0$ \\
 2: & \textbf{repeat} \\
 3: & \hspace*{0.8em}$o\leftarrow\mathrm{Adapt}(r,q)$ \\
 4: & \hspace*{0.8em}$\begin{aligned}[t]
      A\leftarrow{}&\mathrm{First}_{T-|Y|}(\\[-2pt]
                   &\mathrm{PolicyStep}(\pi,o,\xi))
      \end{aligned}$ \\
 5: & \hspace*{0.8em}\textbf{for each} $S\in\mathrm{Split}_{8}(A)$ \textbf{do} \\
 6: & \hspace*{1.6em}$C\leftarrow\mathrm{Render}(q,S)$ \\
 7: & \hspace*{1.6em}$\widehat O\leftarrow\mathrm{WM}(o^{\mathrm{cam}}_0,r,C,\xi)$ \\
 8: & \hspace*{1.6em}$Y\leftarrow Y\mathbin{\|}\widehat O$ \\
 9: & \hspace*{1.6em}$q\leftarrow\mathrm{Last}(S)$ \\
10: & \hspace*{1.6em}$r\leftarrow\mathrm{Last}(\widehat O)$ \\
11: & \hspace*{0.8em}\textbf{end for} \\
12: & \textbf{until} $|Y|=T$ \\
13: & \textbf{return} $Y$
\end{tabular}
\vspace{0.5ex}\hrule height 0.35pt
\end{minipage}
\vspace{-\intextsep}
\end{wrapfigure}

Algorithm~\ref{alg:policy_wm_rollout} gives the closed-loop schedule. At 10\,Hz, FM and ACT each predict an action chunk spanning $H=32$ future time steps, while $\pi_{0.5}$ predicts $H=16$. Each step supplies 14 dual-arm joint targets, so the executed chunk has shape $H\times14$. The latter policy generates actions in a 32D internal space, but only the first 14 dimensions of each step are sent to the robot. The world model consumes the resulting 14D commands in four or two segments of eight actions, respectively.

Here $\mathrm{PolicyStep}(\pi,o,\xi)$ denotes $\pi_{\mathrm{task}}(o)$ for task-specific FM and ACT and $\pi(o,\ell)$ for prompt-conditioned $\pi_{0.5}$, where $\ell$ is the language component of $\xi$. The task prompt represented by $\xi$ is also supplied to the world model's text-conditioning branch for every policy family.

For each segment, we render the robot state and commands into view-aligned action images, then predict eight multimodal frames. The last frame and command update the reference and robot state. The policy replans only after the entire chunk has been executed.

The three RGB views in the real initial observation form the fixed episode anchor, while the full initial observation initializes the current multimodal reference $r$. Thereafter, only $r$ advances with each world-model prediction. The observation adapter extracts the generated RGB and tactile signals required by the policy and attaches the updated robot state. For $\pi_{0.5}$, it additionally attaches the language instruction from $\xi$; task-specific FM and ACT receive no language input. Independently, the task prompt remains available to the world model's text-conditioning branch throughout the rollout. Repeating this cycle produces a full multimodal trajectory without conditioning on recorded future observations.

Here $o_0$ and $q_0$ are the synchronized real initial observation and robot state, $o^{\mathrm{cam}}_0$ is the fixed three-view RGB anchor, $r$ is the current multimodal reference, $\xi$ is the task specification, and $T$ is the requested horizon. In the algorithm, $\operatorname{Render}$ creates the action images, $\operatorname{WM}$ generates the next segment, and $\operatorname{Split}_{8}$ partitions a chunk into segments of at most eight commands. The final chunk is truncated at $T$.

\FloatBarrier
\section{Policy Reinforcement Learning in the World Model}
\label{app:policy_grpo}

\subsection{RL in World Model}
\label{app:grpo_framework}

We optimize a policy by treating the frozen world model as the interaction environment and the frozen State Reward Model as the outcome evaluator.  Starting from one real reset observation, the policy samples multiple action trajectories.  The world model converts each trajectory into a closed-loop multimodal rollout, and the State Reward Model evaluates the generated three-view RGB observations.  Trajectory rewards are normalized only against alternatives generated from the same initial condition.  The resulting group-relative advantages supervise a clipped policy update; gradients update the policy while the world model and State Reward Model remain fixed.

\begin{figure}[H]
\centering
\includegraphics[width=\linewidth]{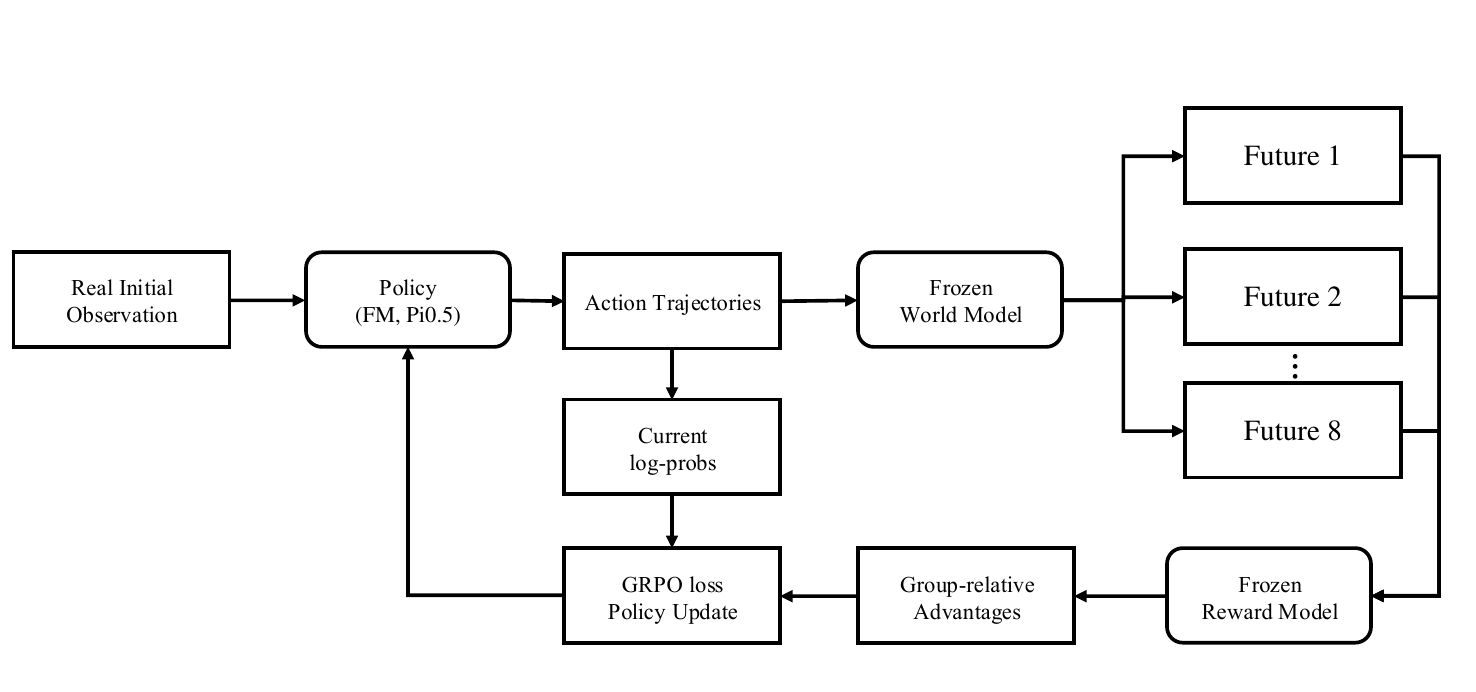}
\caption{\textbf{Policy GRPO in \method{}.} Eight sampled trajectories share one real reset condition; three representative futures are shown.  The frozen world and reward models provide group-relative advantages, while replayed actions provide current policy log-probabilities for the clipped update.}
\label{fig:policy_grpo_pipeline}
\end{figure}

Figure~\ref{fig:policy_grpo_pipeline} summarizes the GRPO sampling process. FM uses 32 integration steps and $\pi_{0.5}$ uses 10. Each policy call selects one step uniformly for a Gaussian Flow-SDE transition; the remaining steps use deterministic Flow-ODE integration. We store the selected index, action latents before and after that transition, current RGB, robot state, and tactile inputs, the tokenized prompt for $\pi_{0.5}$, and the old-policy log-probability. During optimization, FM recomputes the selected Flow-SDE transition, whereas $\pi_{0.5}$ recomputes the selected Flow-SDE denoising transition. The resulting current-policy log-probability forms the GRPO likelihood ratio. The generated rollout supplies the reward and the stored transition supplies actor gradients without differentiating through the world model.

\subsection{Groups, rewards, and objective}
\label{app:grpo_objective}

Each initial condition defines a GRPO group of eight candidate trajectories.  A rollout epoch contains 128 trajectories, or 16 groups, and four rollout epochs are accumulated for one update.  Thus, one update collects 512 trajectories from 64 initial-condition groups.  Candidates within a group share the reset observation and task specification but differ in the stochastic policy transition used to generate their actions.  Rollouts run in the multimodal closed loop described in Appendix~\ref{app:policy_wm_rollout}, with early termination when the generated task outcome satisfies the task criterion.

For candidate trajectory $i$, the State Reward Model evaluates the final frame of the generated chunk and outputs a success probability $p_i^{\mathrm{end}}$.  We use the success of this last frame as the binary trajectory reward,
\begin{equation}
 R_i=\mathbf 1[p_i^{\mathrm{end}}\ge 0.7]\in\{0,1\}.
 \label{eq:appendix_grpo_reward}
\end{equation}
The State Reward Model's continuous confidence therefore determines the terminal success label, while GRPO is optimized with the resulting discrete outcome.

For group $G$, the trajectory advantage is
\begin{equation}
 A_i=\frac{R_i-\mu_G}{\sigma_G+10^{-6}},\qquad
 \mu_G=\frac{1}{|G|}\sum_{j\in G}R_j.
 \label{eq:appendix_grpo_advantage}
\end{equation}
The same $A_i$ is broadcast to all valid action log-probabilities in trajectory $i$, while positions after termination are masked.  Groups in which all eight trajectories succeed or all eight fail contain no relative preference signal and are removed.  Equivalently, for group size eight, training retains groups containing one to seven successful candidates.

Let $\log\pi_{\mathrm{old}}$ be the log-probability recorded during rollout and $\log\pi_\theta$ the value recomputed by the current actor for the same sampled transition.  We form
\begin{equation}
 \rho_{i,k}=\exp\!\left(\log\pi_\theta(a_{i,k}\mid o_{i,k})-
 \log\pi_{\mathrm{old}}(a_{i,k}\mid o_{i,k})\right).
 \label{eq:appendix_grpo_ratio}
\end{equation}
We first apply asymmetric PPO ratio clipping,
\begin{equation}
 \ell^{\mathrm{clip}}_{i,k}
 =\max\!\left(
 -A_i\rho_{i,k},
 -A_i\operatorname{clip}(\rho_{i,k},0.8,1.28)
 \right).
 \label{eq:appendix_grpo_clip_loss}
\end{equation}
We additionally use dual clipping with coefficient $c=3.0$:
\begin{equation}
 \ell^{\mathrm{dual}}_{i,k}
 =\begin{cases}
 \min\!\left(\ell^{\mathrm{clip}}_{i,k},-cA_i\right), & A_i<0,\\[4pt]
 \ell^{\mathrm{clip}}_{i,k}, & A_i\ge 0,
 \end{cases}
 \qquad c=3.0.
 \label{eq:appendix_grpo_dual_loss}
\end{equation}
The final actor loss is
\begin{equation}
 \mathcal L_{\mathrm{actor}}
 =\underset{i,k\in\mathcal V_i}{\operatorname{mean}}
 \ell^{\mathrm{dual}}_{i,k},
 \label{eq:appendix_grpo_loss}
\end{equation}
where $\mathcal V_i$ denotes action log-probability positions that remain valid before trajectory termination. We first apply asymmetric PPO ratio clipping over $[0.8,1.28]$. For trajectories with negative advantage, dual clipping caps the per-position loss at $-cA_i=c|A_i|$, preventing an excessively large update when the current policy assigns a much higher probability to an action sampled from a below-average trajectory. For non-negative advantages, the dual-clip term has no effect. FM and $\pi_{0.5}$ differ in how the stochastic Flow-SDE transition and its log-probability are constructed, but both policies are optimized with this same dual-clipped GRPO actor objective. The old policy is the snapshot that generated the current rollout batch, not the original supervised checkpoint. We do not train a critic and do not add KL, entropy, or value losses in the reported configuration.

\begin{table}[H]
\centering
\small
\caption{\textbf{Common GRPO configuration for FM and $\pi_{0.5}$.}}
\label{tab:grpo_common_configuration}
\begin{tabularx}{0.94\linewidth}{@{}lXlX@{}}
\toprule
\textbf{Parameter} & \textbf{Value} & \textbf{Parameter} & \textbf{Value} \\
\midrule
Group size & 8 trajectories & Actor global batch & 128 \\
Rollout epochs/update & 4 & Trajectories/update & 512 \\
Micro batch & 8 & Actor update epochs & 1 \\
Learning rate & $5\times10^{-6}$ & Adam $(\beta_1,\beta_2)$ & $(0.9,0.95)$ \\
Adam $\epsilon$ & $10^{-8}$ & Weight decay & $0.01$ \\
Gradient clipping & $1.0$ & Ratio clipping & $[0.8,1.28]$ \\
Dual clip & $3.0$ & KL / entropy coefficient & $0.0/0.0$ \\
Critic & disabled & Checkpoint interval & 4 updates \\
Compute & 2 nodes $\times$ 8 GPUs & Reward threshold & $0.7$ \\
\bottomrule
\end{tabularx}
\end{table}

\section{Metric Computation}
\label{app:metric_definitions}
This section defines how each metric in Tables~\ref{tab:baselines} and~\ref{tab:ablations} is computed. Ground-truth and predicted quantities are denoted without and with a hat, respectively. Higher PSNR, SSIM, consistency scores, and F1 indicate better performance; lower LPIPS, distributional distances, tracking errors, timing errors, and MAEs are better.

\paragraph{Image reconstruction and distributional distances.}
For an aligned RGB frame pair $I,\hat I\in[0,1]^{H\times W\times3}$, the mean-squared error and peak signal-to-noise ratio are
\begin{equation}
 e_{\mathrm{MSE}}=\frac{\|I-\hat I\|_F^2}{3HW},
 \qquad \mathrm{PSNR}=-10\log_{10}e_{\mathrm{MSE}}.
\end{equation}
PSNR is computed for each frame and then averaged. Structural similarity is computed over corresponding local windows as
\begin{equation}
 \mathrm{SSIM}(u,v)=
 \frac{(2\mu_u\mu_v+C_1)(2\sigma_{uv}+C_2)}
 {(\mu_u^2+\mu_v^2+C_1)(\sigma_u^2+\sigma_v^2+C_2)},
\end{equation}
where $\mu$, $\sigma^2$, and $\sigma_{uv}$ denote local means, variances, and covariance, and $C_1=(0.01L)^2$, $C_2=(0.03L)^2$ for intensity range $L$. LPIPS is the spatially averaged, learned weighted squared distance between normalized intermediate AlexNet features of $I$ and $\hat I$, summed across feature layers.

FID and FVD compare Gaussian feature distributions of ground-truth and generated data using the Fr\'echet distance
\begin{equation}
 D=\|\mu_g-\mu_p\|_2^2+
 \operatorname{Tr}\!\left(\Sigma_g+\Sigma_p
 -2(\Sigma_g^{1/2}\Sigma_p\Sigma_g^{1/2})^{1/2}\right).
\end{equation}
FID applies this equation to image-level Inception features, whereas FVD applies it to I3D features extracted from temporally ordered video clips.

\paragraph{Temporal consistency.}
Subject and background consistency use normalized frame features $z_t$ from DINO and CLIP, respectively. Each frame is compared with both the first and preceding frames:
\begin{equation}
 C=\frac{100}{2(T-1)}\sum_{t=1}^{T-1}
 \left[\max(0,z_t^\top z_0)+\max(0,z_t^\top z_{t-1})\right].
\end{equation}
Motion smoothness reconstructs an intermediate frame from its two neighbors with a pretrained interpolation model and computes
\begin{equation}
 C_{\mathrm{smooth}}=100\left(1-
 \operatorname{MAE}(I_{\mathrm{mid}},I_{\mathrm{interp}})\right),
\end{equation}
averaged over all reconstructed intermediate frames.

\paragraph{End-effector and object tracking.}
Let $p_t$ and $\hat p_t$ be corresponding two-dimensional positions in the ground-truth and generated videos. End-effector position is the mean position of the gripper-jaw points, while object position is obtained from the object track. With visibility indicator $v_t$, the average tracking error is
\begin{equation}
 E_{\mathrm{track}}=
 \frac{\sum_{t=1}^{T-1}v_t\|\hat p_t-p_t\|_2}
 {\sum_{t=1}^{T-1}v_t},
\end{equation}
and is reported in pixels.

\paragraph{Tactile activity and contact error.}
For one tactile sensor, let $I_t$ and $\hat I_t$ denote the ground-truth and predicted difference images and let $R=I_0$ be the reference image. Pixel activity is
\begin{equation}
 d_t(x)=\tfrac13\|I_t(x)-R(x)\|_1,\qquad
 A_t(x)=\mathbf1[d_t(x)>\tau_d],\qquad
 \hat A_t(x)=\mathbf1[\tfrac13\|\hat I_t(x)-R(x)\|_1>\tau_d],
\end{equation}
where $\tau_d$ is the fixed image-change threshold. Contact MAE measures RGB prediction error over ground-truth active pixels:
\begin{equation}
 \mathrm{Contact\ MAE}=
 \frac{\sum_{t,x}A_t(x)\|I_t(x)-\hat I_t(x)\|_1}
 {3\sum_{t,x}A_t(x)}.
\end{equation}

\paragraph{Marker-displacement errors.}
For each valid marker--time pair $i$, let $f_i,\hat f_i\in\mathbb{R}^3$ be the ground-truth and predicted displacement vectors. Define the active and idle sets as
\begin{equation}
 \mathcal A_f=\{i:\|f_i\|_2>\tau_f\},\qquad
 \mathcal Z_f=\{i:\|f_i\|_2\leq\tau_f\},
\end{equation}
where $\tau_f$ is the fixed displacement threshold. The three marker metrics are
\begin{align}
 \mathrm{Active\ MAE}&=\frac{1}{3|\mathcal A_f|}
 \sum_{i\in\mathcal A_f}\|\hat f_i-f_i\|_1,\\
 \text{Idle FP-MAE}&=\frac{1}{3|\mathcal Z_f|}
 \sum_{i\in\mathcal Z_f}\|\hat f_i\|_1,\\
 \mathrm{Amp.\ MAE}&=\frac{1}{|\mathcal A_f|}
 \sum_{i\in\mathcal A_f}
 \left|\|\hat f_i\|_2-\|f_i\|_2\right|.
\end{align}
Active MAE compares displacement components, Idle FP-MAE measures spurious motion on idle markers, and Amp. MAE compares displacement magnitudes. The reported values are multiplied by $10^3$.

\paragraph{Event F1 and boundary timing.}
For each tactile sensor, a frame is active when its active-pixel fraction exceeds a fixed event threshold. The same rule is applied to ground truth and prediction. Inactive gaps of at most one frame are filled, active runs shorter than two frames are removed, and each remaining run forms an interval $[a,b)$. Ground-truth and predicted intervals are matched one-to-one when their temporal intersection-over-union is at least $0.1$.

If $K$ pairs are matched among $N_g$ ground-truth and $N_p$ predicted events, then
\begin{equation}
 \mathrm{Precision}=\frac{K}{N_p},\qquad
 \mathrm{Recall}=\frac{K}{N_g},\qquad
 \mathrm{F1}=\frac{2K}{N_g+N_p}.
\end{equation}
For each matched pair, onset and offset errors are
\begin{equation}
 E_{\mathrm{onset}}=\frac{1}{K}\sum_{j=1}^{K}
 \frac{|a_j^{g}-a_j^{p}|}{r},\qquad
 E_{\mathrm{offset}}=\frac{1}{K}\sum_{j=1}^{K}
 \frac{|b_j^{g}-b_j^{p}|}{r},
\end{equation}
where $r$ is the frame rate, so both errors are reported in seconds.

\paragraph{Aggregation.}
Framewise image and temporal scores are averaged over time. Pixel and marker errors pool their eligible elements within each tactile sensor before averaging across sensors and trajectories. Event F1 and timing errors are computed per sensor trajectory and then macro-averaged. FID and FVD are computed directly from the corresponding collections of image and video features.

\FloatBarrier
\section{Visualization Analysis}
\label{app:visualization_analysis}

\begingroup
\setlength{\intextsep}{4pt}
\setlength{\textfloatsep}{4pt}
\setlength{\abovecaptionskip}{2pt}
\setlength{\belowcaptionskip}{2pt}

This section presents qualitative comparisons with video-generation baselines and examines how multimodal attention changes across network depth. The displayed cases are diagnostic examples rather than estimates of failure frequency.

\subsection{Qualitative Comparison with Baselines}
\label{app:baseline_comparison}

\begin{figure}[H]
  \centering
  \includegraphics[width=0.94\linewidth]{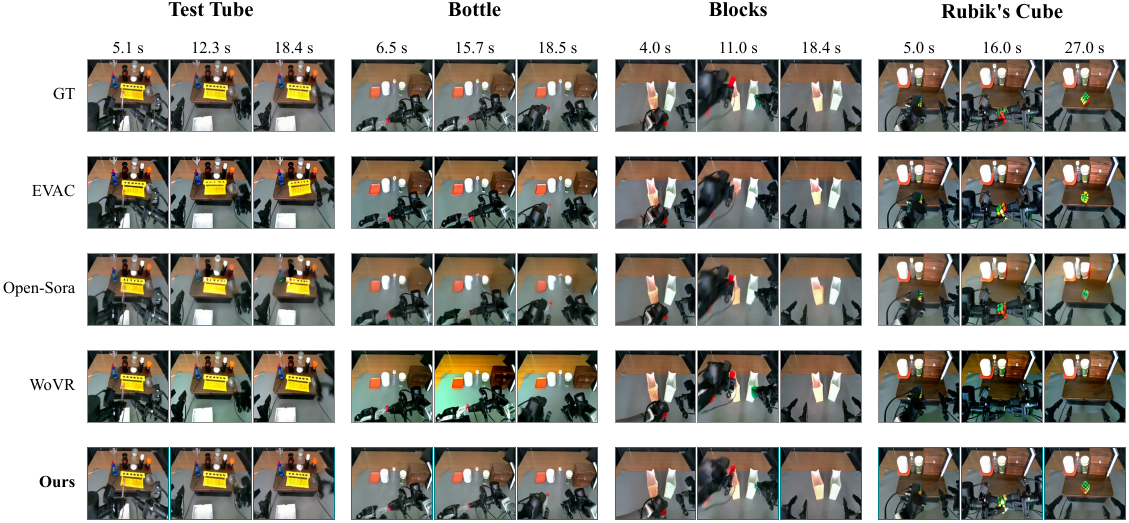}
  \caption{\textbf{Qualitative comparison across four tasks.} GT and model rollouts are shown at aligned timestamps.}
  \label{fig:baseline_comparison}
\end{figure}

Figure~\ref{fig:baseline_comparison} shows that the displayed \method{} rollouts retain small objects and scene layout more consistently. Open-Sora smooths small container boundaries, WoVR introduces a strong appearance shift, and several baseline predictions alter the bag openings. These selected examples are descriptive rather than estimates of failure frequency.

\vspace{-0.35em}
\subsection{Layer-Wise Multimodal Attention Analysis}
\label{app:attention_analysis}

We visualize modality-level attention for Bottle case~01, averaged over all heads and five denoising steps at layers 0, 14, and 29. Rows denote query modalities and columns denote key modalities; uniform attention would allocate one third of the mass to each equally sized modality band.

\begin{figure}[H]
  \centering
  \includegraphics[width=0.88\linewidth]{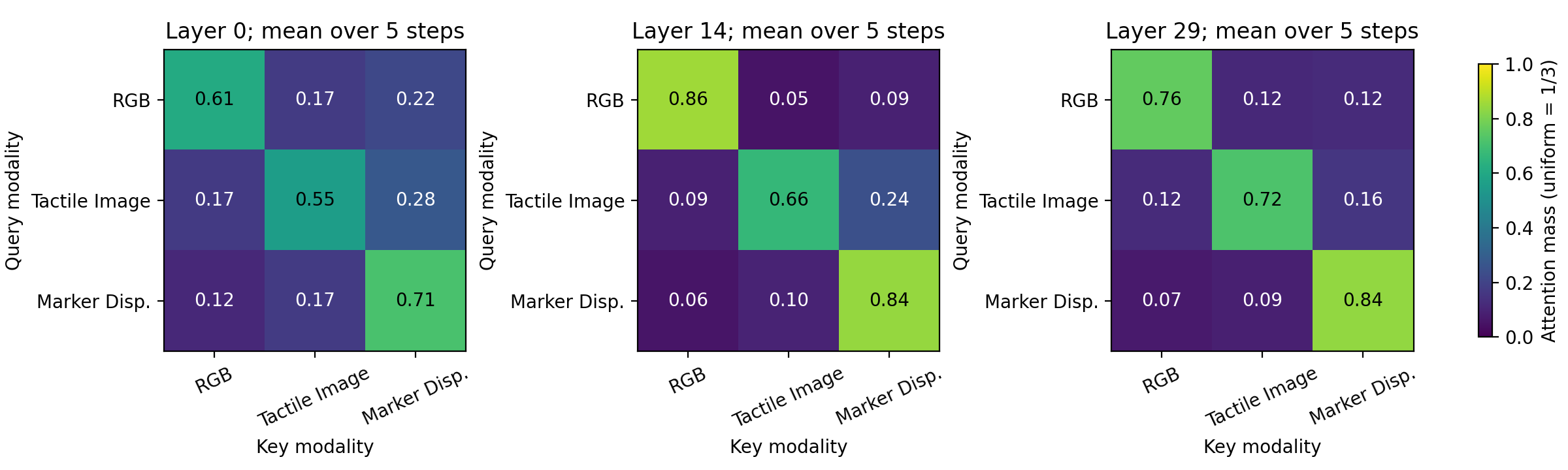}
  \caption{\textbf{Layer-wise multimodal attention for Bottle case~01.} Off-diagonal entries show cross-modal routing, while diagonal entries show within-modality attention.}
  \label{fig:attention_layerwise}
\end{figure}

Figure~\ref{fig:attention_layerwise} shows the strongest cross-modal exchange in the shallow layer: at layer~0, RGB, tactile-image, and marker-displacement queries allocate 0.39, 0.45, and 0.29 of their attention outside their own modality. The diagonal mass becomes dominant at layers~14 and 29, where the three within-modality entries are $(0.86, 0.66, 0.84)$ and $(0.76, 0.72, 0.84)$, respectively. We find that modality fusion occurs primarily in the initial attention blocks; after cross-modal attention in the shallow representation space, deeper blocks concentrate increasingly on their respective modalities.

\vspace{-0.25em}
\noindent The following six pages present qualitative rollout comparisons. The first four compare ground-truth (GT) rollouts with \method{} predictions (Pred) conditioned on the same action sequence. The final two compare real-world executions of policies before and after post-training: Base denotes the original supervised fine-tuned policy, whereas GRPO denotes the corresponding policy after Group Relative Policy Optimization through imagined interactions in \method{}. In all comparisons, frames progress from left to right.

\endgroup

\clearpage
\iclrfinaltrue
\thispagestyle{empty}
\AddToShipoutPictureFG*{%
  \AtPageCenter{%
    \makebox(0,0){%
      \includegraphics[page=1,height=\paperheight,keepaspectratio]{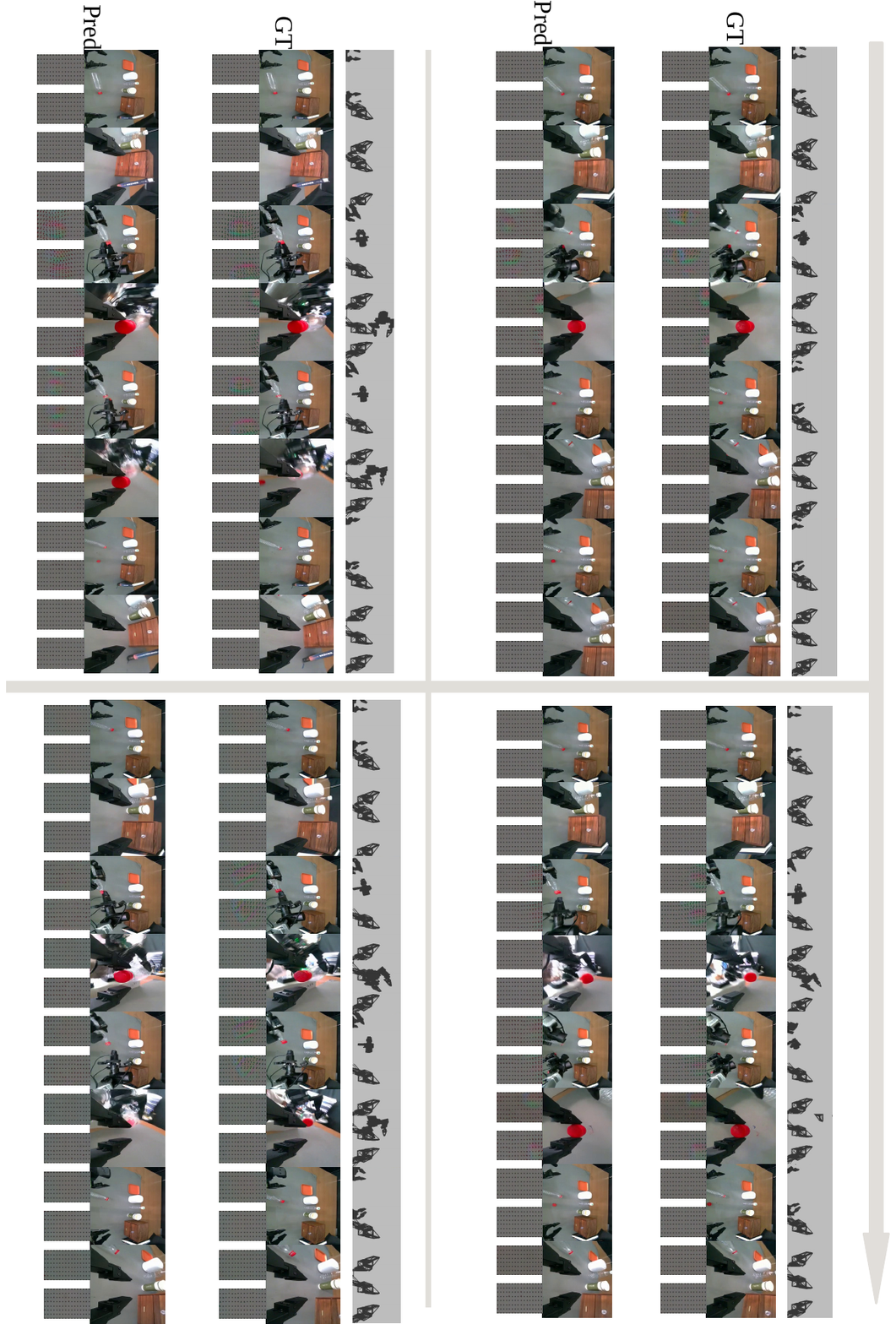}%
    }%
  }%
}
\null
\clearpage
\thispagestyle{empty}
\AddToShipoutPictureFG*{%
  \AtPageCenter{%
    \makebox(0,0){%
      \includegraphics[page=2,height=\paperheight,keepaspectratio]{assets/analysis/appendix_qualitative_cases_2page_A4_portrait_v4_no_state.pdf}%
    }%
  }%
}
\null
\clearpage
\thispagestyle{empty}
\AddToShipoutPictureFG*{%
  \AtPageCenter{%
    \makebox(0,0){%
      \includegraphics[page=3,height=\paperheight,keepaspectratio]{assets/analysis/appendix_qualitative_cases_2page_A4_portrait_v4_no_state.pdf}%
    }%
  }%
}
\null
\clearpage
\thispagestyle{empty}
\AddToShipoutPictureFG*{%
  \AtPageCenter{%
    \makebox(0,0){%
      \includegraphics[page=4,height=\paperheight,keepaspectratio]{assets/analysis/appendix_qualitative_cases_2page_A4_portrait_v4_no_state.pdf}%
    }%
  }%
}
\null
\iclrfinalfalse

\FloatBarrier
\clearpage
\iclrfinaltrue
\thispagestyle{empty}
\AddToShipoutPictureFG*{%
  \AtPageCenter{%
    \makebox(0,0){%
      \includegraphics[page=1,height=\paperheight,keepaspectratio]{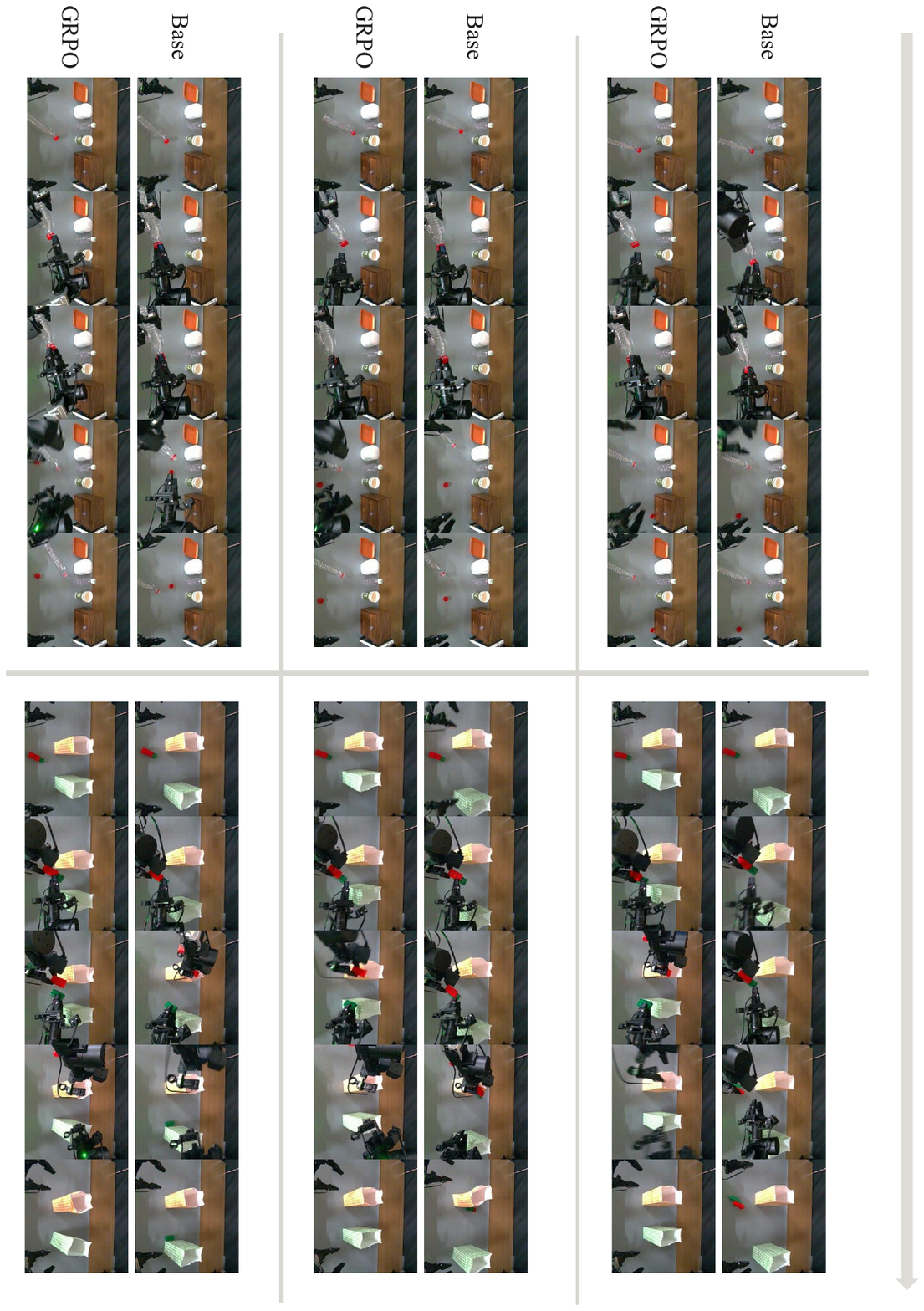}%
    }%
  }%
}
\null
\clearpage
\thispagestyle{empty}
\AddToShipoutPictureFG*{%
  \AtPageCenter{%
    \makebox(0,0){%
      \includegraphics[page=2,height=\paperheight,keepaspectratio]{assets/analysis/GRPO_learning.pdf}%
    }%
  }%
}
\null
\iclrfinalfalse

\end{document}